%% file: mia.tex
\documentclass{article}
\usepackage[accepted]{icml2022}
\icmltitlerunning{Almost for Free: Crafting Adversarial Examples with Convolutional Image Filters}

\usepackage{microtype}
\usepackage{graphicx}
\usepackage{subfigure}
\usepackage{booktabs} % for professional tables
\usepackage{numprint}
\usepackage{multirow}

\usepackage{hyperref}

\usepackage{amsmath}
\usepackage{amssymb}
\usepackage{mathtools}
\usepackage{amsthm}
\usepackage{xspace}

\usepackage{tikz}
\usepackage{pgfplots}
\pgfplotsset{compat=1.8}
\usetikzlibrary{positioning, backgrounds, arrows,fit,matrix, arrows.meta,calc,shapes.geometric, calc, patterns}
\usepgfplotslibrary{groupplots}
\usepgfplotslibrary{fillbetween}
\usepgfplotslibrary{statistics}

\usepackage[capitalize,noabbrev]{cleveref}

\theoremstyle{plain}

\theoremstyle{definition}

\theoremstyle{remark}

\newcommand{\vecx}{\mathbf{x}\xspace}
\newcommand{\vecdelta}{\mathbf{\delta}\xspace}
\newcommand{\vectheta}{\mathbf{\theta}\xspace}

\DeclareMathOperator*{\argmin}{argmin\,}

\AtBeginDocument{%
  \providecommand\BibTeX{{%
    \normalfont B\kern-0.5em{\scshape i\kern-0.25em b}\kern-0.8em\TeX}}}

\begin{document}
\renewcommand{\Notice@String}{}
\gdef\icmlcorrespondingauthor@text{}
\makeatletter
\renewcommand{\printAffiliationsAndNotice}[1]{%
\stepcounter{@affiliationcounter}%
{\let\thefootnote\relax\footnotetext{\hspace*{-\footnotesep}%
\forloop{@affilnum}{1}{\value{@affilnum} < \value{@affiliationcounter}}{
\textsuperscript{\arabic{@affilnum}}\ifcsname @affilname\the@affilnum\endcsname%
\csname @affilname\the@affilnum\endcsname%
\else
{\bf AUTHORERR: Missing \textbackslash{}icmlaffiliation.}
\fi
}}}}
\makeatother

\twocolumn[
\icmltitle{Almost for Free: Crafting Adversarial Examples \\ with
Convolutional Image Filters}

\begin{icmlauthorlist}
\icmlauthor{Alexander Warnecke}{uni1}
\icmlauthor{Konrad Rieck}{uni1}
\end{icmlauthorlist}

\icmlaffiliation{uni1}{BIFOLD \& TU Berlin}

\vskip 0.3in]
\printAffiliationsAndNotice{}

\begin{abstract}
Adversarial examples in machine learning are typically generated using gradients, obtained either directly through access to the  model or approximated via  queries to it. 
In this paper, we propose a much simpler approach to craft adversarial examples, drawing inspiration from insights of explainable machine learning.
In particular, we design \emph{adversarial image filters} that are based on classic edge detection algorithms but optimized to deceive learning models.
The resulting untargeted attacks are transferable and require only a single pass over the input.
Empirically, we find that $3 \times 3$ filters already enable  success rates between 30\% and 80\% on different neural networks. Compared to related approaches using generative models for crafting adversarial examples, we reduce the number of parameters by five orders of magnitude, resulting in a very efficient attack.
When investigating the parameters of the learned filters, we observe interesting properties such as a high transferability between models and structures common to classic image filters. Our results provide further insights into the vulnerability of neural networks and their fragility to malicious noise.
\end{abstract}

\vspace{3mm}
\section{Introduction}
\label{sect:intro}

Deep neural networks achieve remarkable performance in various learning tasks, particularly in the vision domain~\cite{KriSutHin12, SimZis15}. These successes, however, are overshadowed by their vulnerability to adversarial examples---deceptive inputs designed to mislead predictions~\citep{SzeZarSut+14}. The prevalent paradigm for constructing these attacks rests on using gradients of the learning model, obtained either through direct access to its parameters~\citep{CarWag17, GooShlSze15} or by approximation through multiple queries to it~\citep{CheZhaSha+17, IlyEngAth+18}.

\begin{figure}[t]
    \hspace{-4mm}
    \input{figures/intro/adv-example-eagle}
    \caption{Example of adversarial image filters. An input image (left) is perturbed using a convolutional filter (middle) to generate an adversarial example (right).}
    \label{fig:intro-figure}
    \vspace{-3mm}
\end{figure}
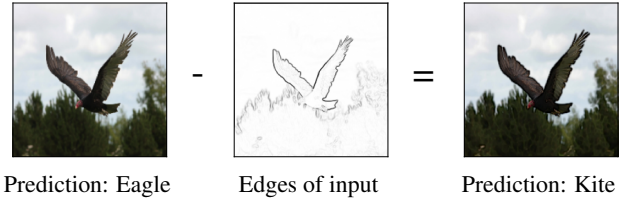

In this paper, we deviate from this paradigm and propose \emph{adversarial image filters} as a simple strategy for crafting adversarial examples. These filters deceive learning models through a single convolution and hence do not require the gradient information. Our approach is inspired by the connection between edges and gradients observed in explainable machine learning~\cite{AdeGilMue+18, AncCeoOtr+18}. Starting from a classic Sobel filter for edge detection \cite{SobFel73}, our attack optimizes the convolution weights of the filter to maximize mispredictions of learning models. The resulting untargeted attack requires only a single pass over the input and transfers well between models.
An example of our attack can be seen in \cref{fig:intro-figure}, where a single convolution induces a misclassification.

We empirically evaluate our attack with different convolutional filter sizes and find that a small $3 \times 3$ filter already achieves attack success rates between 45\% and 94\%  against learning models such as ResNet50~\cite{HeZhaRenSun16}. The visual changes are hardly noticeable, remaining above a PSNR of $20$.
In comparison to related approaches that use generative models to create adversarial examples~\citep{SeyAlhFaw+16, BalFis18, XiaLiZhu+18}, our attack reduces the number of parameters by five orders of magnitude. This substantial reduction not only enhances efficiency but drastically simplifies the overall attack mechanism.
Furthermore, the learned filters demonstrate better performance when compared to attacks based on conventional image filters \citep{AgaVatSin+20, GuoXuXie+20}. While blurring filters can naturally induce mispredictions, they cannot achieve similar attack performance at the same PSNR level.

When investigating the parameters of the learned $3 \times 3$ filters, we observe interesting properties: the filters are characterized by a high level of transferability across models. Due to their small size, model-specific properties can hardly be encoded in their weights, resulting in a largely model-independent attack. While the learned weight structure shares similarities with existing filters, such as the Laplacian of Gaussian filter, its specific configuration is unique and forms a new filter type. This filter type offers new view on  the fragility of neural networks to malicious noise. By perturbing image regions with a single convolution, we achieve a notable decline in model performance, suggesting that complex approaches are not always necessary for effective attacks.

It is clear that a simple convolutional filter cannot compete with sophisticated optimization strategies for crafting adversarial examples. Still, our attack can serve as a lower bound on the performance of untargeted attacks. The run-time and memory complexity of more advanced strategies need to be weighed against our filters to put their performance improvements into perspective.

%The majority of work that exists in this field either requires tens of thousands of input examples to learn a generative model~\cite{SeyAlhFaw+16} or uses
%model comprising millions of parameters to craft perturbations for the ImageNet dataset~\cite{BalFis18, XiaLiZhu+18}.

%By leveraging the connection between
%saliency maps and edges~\cite{AdeGilMue+18} as well as the similarity between the gradient and many explanation methods~\cite{AncCeoOtr+18}, we highlight a new connection showing the sensitivity of neural networks to edges and structure and show that a simple
%Sobel filter for edge detection\cite{SobFel73} comprising $3\times 3$ parameters can be used to craft untargeted attacks  completely independent from the model or its parameters.

%While these insights have been used to fool networks, for example using
%blurring filters~\citep{AgaVatSin+20, GuoXuXie+20} we ask the question
%how a convolutionl filter optimized to generate adversarial examples by
%sliding it over the image would look like. 

%To this end, we introduce
%the concept of \emph{adversarial filters}, an adaptive attack where we optimize the parameters of a convolutional filter using a small fraction of example images.

\section{Related Work}
The phenomenon of adversarial examples (AEs) in deep learning was first discovered by~\citet{SzeZarSut+14} and has since become an active research area, resembling a cat-and-mouse game where new defenses are swiftly proposed and then quickly broken.
In the case where model parameters are available to the adversary (white-box setup), AEs can be constructed using gradient information~\citep[e.g.][]{CarWag17,GooShlSze15,PapMcdGoo+16,MooFawFro16}. In the case where parameters are unavailable (black-box setup), various gradient approximations can be employed, such as shadow models~\cite{PapMcDJha+16}, finite differences~\citep{CheZhaSha+17}, Gaussian noise averages~\citep{IlyEngAth+18}, evolution strategies~\citep{IlyEngMad19}, and random navigations through the input space~\citep{GuoGarYou+19}. Even when only the network’s decision, rather than softmax scores, is available, strategies for crafting adversarial examples are available~\citep{CheJor19, BreRauBet17}.

There are only a few approaches to crafting AEs that fall outside these two  attack scenarios. \citet{SeyAlhFaw+16} introduce \emph{universal adversarial perturbations}, which involve a \emph{single} perturbation vector that can be added to \emph{any} image to cause a misclassification. Similarly, \citet{BalFis18} train a neural network to generate an adversarial perturbation for a given input using an adversarial criterion in the loss function. This approach shares similarities with adversarial filters, yet the trained networks require between $3.4$ and $233.7$ million parameters for a successful attack. 
The underlying approach has been further adapted to generative models~\cite{XiaLiZhu+18, PouKatGao+18}, most prominently generative adversarial networks (GANs)~\cite{GooPouMir+14}, which transform a random noise vector into an adversarial perturbation. As part of this transformation, properties like transferability can be directly incorporated into the learning problem~\cite{NakSal21}. However, the resulting architectures are complex, again involving a large number of model parameters.

Closest to our work are attacks that rely on image transformations, leveraging the sensitivity of neural networks to edges and structures. For example, \citet{GeiRubMich+18} maliciously replace textures in images. Similarly, \citet{AgaVatSin+20} and \citet{GuoXuXie+20} discuss the use of image filters for blur and motion blur to craft adversarial examples. \citet{ShaShaChan+20} propose a special type of adversarial example where contours are enhanced by decomposing the image into structural and residual parts using a neural network. \citet{HeAiLei+23} show that the robustness of neural networks increases when using these examples for adversarial training.
While our approach is related to this work, the concept of training an adversarial filter has not been explored so far. As we discuss in the evaluation, better performance can be attained if the convolution is specifically adapted to this malicious task.

% \citet{AgaRatVat+22} \citet{GavKeu23} ?

\section{Motivation and Background}

We begin by defining the notation used throughout the paper and motivating our approach to generating adversarial examples with convolutional filters.

%As we focus on images, we assume that  $d = w \times h \times 3$, where $w$ is the width, $h$ is the height, and there are three color channels.

Let $\vectheta \in \mathbb{R}^m$ represent a learning model with $m$ parameters, and let $f_\vectheta$ denote the associated prediction function. For an input $\vecx \in \mathbb{R}^d$, the model yields a prediction $f_\vectheta(\vecx) \in \{1, \ldots, c\}$, where $c$ denotes the number of classes. Given a labeled dataset $D$ consisting of training examples, the optimal model parameters $\vectheta^*$ are determined by minimizing the empirical risk as follows
\begin{equation}
\label{eq:empirical-risk}
    \vectheta^* = \argmin_{\vectheta\in\mathbb{R}^m} \sum_{(\vecx,y)\in D} \ell(\vecx,y,\vectheta),
\end{equation}
where $\ell(\vecx, y, \vectheta)$ is a loss function that measures the difference between a prediction $f_\vectheta(\vecx)$ and the true label $y$. In the remainder of this paper
we use public models that represent solutions of \cref{eq:empirical-risk} for benchmark datasets.

An untargeted adversarial example is an input $\vecx'=\vecx + \vecdelta$ that results from adding an imperceptible perturbation $\vecdelta\in\mathbb{R}^d$ to $\vecx$ and
has the property that $f_\vectheta(\vecx)\neq f_\vectheta(\vecx')$~\cite{SzeZarSut+14}. Generating minimal perturbations invisible for the human eye can be defined as
solutions of the optimization problem
\begin{equation}
	\label{eq:chap2-adversarial-example}
	\min_{\mathbf{\delta}\in\mathbb{R}^d} \lVert\mathbf{\delta}\rVert_p\quad\text{s.t.}\quad f_\theta(\vecx+\mathbf{\delta}) \neq f_\theta(\vecx),
\end{equation}
for some $L^p$ norm, $p\in\{1,2,\infty\}$.
If we denote by $\mathcal{A}$ an algorithm that generates adversarial examples, existing approaches can be broadly categorized using the function signatures of $\mathcal{A}$:

\paragraph{White-box attacks}
If the model parameters are available, $\mathcal{A}$ is given by $\vecdelta = \mathcal{A}(\vecx, \theta)$, that is, we know the model architecture and its
parameters $\theta$. An example for such an approach is the FGSM algorithm~\cite{GooShlSze15} that leverages first-order optimization techniques and solves \cref{eq:chap2-adversarial-example} by performing update steps of the form
\begin{equation}
\label{eq:chap2-fgsm}
\vecx^{(t+1)} = \vecx^{(t)} + \tau \cdot \text{sgn}\big(\nabla_\vecx \ell(\vecx^{(t)}, f_\theta(\vecx^{(t)}), \theta)\big),
\end{equation}
for a loss function $\ell$, a small learning rate $\tau>0$ and a data point $\vecx = \vecx^{(0)}$. The advantage of white-box algorithms is the high quality of the outcome and the fact that an AE can almost always be found. On the other hand, access to the model architecture and parameters is a strong assumption for real-world attacks.

\paragraph{Black-box attacks} Generating adversarial examples in a black-box setting corresponds to the function signature $\vecdelta = \mathcal{A}(\vecx, f_\vectheta)$, where $f_\vectheta$ is treated as an \emph{oracle} function where we neither know the architecture, nor the parameters of the underlying model. Solutions to this approach usually try to approximate white-box algorithms by estimating the gradient $\nabla_{\vecx}f$, which constitutes the
direction into which we have to move the input $\vecx$. A straight-forward approach is given by a finite difference estimation given by
\begin{equation}
\label{eq:chap2-zoo}
\frac{\partial f_\theta(\vecx)}{\partial x_j} \approx \frac{f_\theta(\mathbf{x}+\beta\mathbf{e}_j)-f_\theta(\mathbf{x}-\beta\mathbf{e}_j)}{2\beta}
\end{equation}
for example where $\beta>0$ is a small step size and $\mathbf{e}_j$ denotes the $j$-th unit vector in $\mathbb{R}^d$~\cite{CheZhaSha+17}. Although black-box algorithms can yield excellent results, they require a lot of queries to the oracle which may be rejected by the application in practice~\citep{DebCarTra24}.

\paragraph{Transformation attacks.} Interestingly, the process of crafting adversarial examples can also be described as a transformation that receives an input and returns its perturbed version. The signature is given by $\vecdelta = \mathcal{A}(\vecx, T)$, where $T:\mathbb{R}^d \rightarrow \mathbb{R}^d$ is a function that maps an input $\vecx$ to its adversarial version $\vecx'$.
In the literature, $T$ has been chosen as either another neural network directly transforming the input \citep{BalFis18} or as a generative adversarial network (GAN) that transforms a random image into a perturbation $\vecdelta$ given $\vecx$ \citep{XiaLiZhu+18}. These approaches are considered gray-box attacks, as $T$ is a learning function with a set of parameters $\vectheta_T$ that must be optimized. Consequently, a dataset $D_T$ from the same distribution as $D$ is required, along with access to the model parameters to incorporate the adversarial criterion into the learning problem.
Once optimized, however, these approaches can generate adversarial examples (AEs) for any input, which is a useful property for concepts like adversarial training \citep{MadMakSch+18}.

In this paper, we propose two transformation attacks related to edge detection from computer vision that employ convolutional filters as basis for the underlying transformation function.

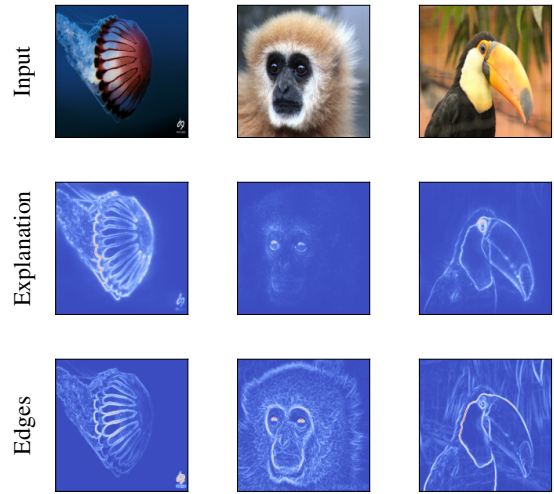
\begin{figure}[h!]
    \centering
    \input{figures/intro-figure-grid/intro-mia}
    \caption[Input images, LRP explanations and edges detected by a Sobel filter]
    {Input images (top row) with LRP explanations (middle row) and edges (bottom row) detected by a Sobel filter.}
    \label{fig:mia-intro-grid}
\end{figure}

\paragraph{Model-independent attacks.} 
For our first new attack, we extend the gray-box assumptions to a case where $T$ is completely independent of the model, that is, it neither uses model parameters nor oracle predictions. Our motivation for constructing adversarial examples in this manner stems from the work of \citet{AdeGilMue+18}. They demonstrate that approaches to explain neural networks through saliency maps \citep{BacBinMon+15, SunTalYan17} share similarities with simple edge detection.
For intuition, \cref{fig:mia-intro-grid} shows explanations from the LPR explanation method \citep{BacBinMon+15} and edges generated by a simple Sobel filter \citep{SobFel73} for inputs to a VGG13 network pre-trained on the ImageNet dataset. It is striking that the information provided by the explanation, although somewhat more nuanced, significantly overlaps with the edges extracted from the image. 

Concurrently, \citet{AncCeoOtr+18} point out that---under certain conditions---explanations are equivalent to the gradient $\nabla_{\vecx} f_\theta(\vecx)$, the core building block of white-box algorithms for crafting adversarial examples. Consequently, we ask: \emph{If edges have similarities with the gradients $\nabla_{\vecx} f_\theta(\vecx)$, could edge detection filters be adapted to construct adversarial examples?}

Combining the observations from above we can define a strategy to craft model independent examples as 
\begin{equation}
\label{eq:mia-general}
    \mathbf{\delta} = \mathcal{A}(\vecx) = \rho(\mathcal{E}(\vecx),
\end{equation}

\noindent where $\mathcal{E}$ is an edge detection method and $\rho$ is a processing function. 
As a simple example, consider the function
\begin{equation}
\label{eq:mia-sobel}
\rho(\mathcal{E}(\vecx)) = \pm\mu\mathcal{E}(\vecx),
\end{equation}

\noindent where $\mu\in\mathbb{R}$ is a scalar. Here, we aim to confuse the
learning model by perturbing the edges in the image. If $\mathcal{E}(\vecx) \approx \nabla_{\vecx}f_\theta(\vecx)$ and $\mu >0$, this rule can be seen as
a single approximation step in the FGSM algorithm~\cite{GooShlSze15}. If $\mu<0$, this procedure simply removes the information contained in the edge pixels by
shifting their value closer to zero. In both cases, it is important to assure that the output of $\rho$ remains a valid image, for example, by applying a suited clipping function to the outcome.

\paragraph{Model-dependent attack}
As a second approach, we return to the original setting of transformation attacks but use convolutional filters as the learning model. Specifically, we optimize their values based on a small training dataset with access to $f_\vectheta$. As we will see in \cref{sect:adaptive-filter-attack}, this approach yields better results compared to the first strategy but produces slightly less imperceptible and less successful perturbations compared to other gray-box strategies. However, due to the small size of the filters, the computational complexity is significantly reduced, which raises questions about the necessity of complex learning models for transformation attacks.

\section{Approach}

\begin{figure*}[t]
    \centering
    \input{figures/blurring-filter}
    \vspace{-6mm}
    \caption{Examples for spatial filters that can be convolved with an image to transform it in various ways.}
    \label{fig:filter-examples}
\end{figure*}

In the following section, we formulate two different versions of our filter-based attack. One that is based on conventional edge filters and one that optimizes
a given filter using some data similar to the training set of the target model, resulting in an adversarial filter. 

Edge detection is a large research field that is also transformed by learning
based approaches~\citep[see e.g.][for an overview]{SunLeiChe+22} but we will investigate implementations of \cref{eq:mia-sobel} using classic approaches from
computer vision that are based on spatial kernels. Some examples for such filters are given in \cref{fig:filter-examples} and illustrate that basic tasks like image
smoothing (Filters (a) and (b)), image sharpening (Filter (c)) and combinations thereof (Filter (d)) can be computed with a single convolution operation. These approaches are light weight, easy to implement and allow for a direct investigation of the outcome.

\subsection{Edge Filter Attack}
\label{sect:static-filter-attack}
We start with an implementation of \cref{eq:mia-sobel} based on a Sobel filter, a classic approach from computer vision to detect edges in images~\cite{SobFel73}. Given a gray-scale image $\vecx\in\mathbb{R}^{w\times h}$, the Sobel filter computes a convolution between $\vecx$ and its spatial filters $K_x$ and $K_y$ given
by
\begin{equation}
\label{eq:sobel-filter-raw}
	K_x = \begin{bmatrix}
		1&2&1\\
		0&0&0\\
		-1&-2&-1
	\end{bmatrix}
    \qquad 
	K_y = \begin{bmatrix}
		1&0&-1\\
		2&0&-2\\
		1&0&-1
	\end{bmatrix}.
\end{equation}

Afterwards, the the gradient magnitude at each pixel is computed by
\begin{equation}
\label{eq:sobel-filter}
	S(\vecx) = \sqrt{\big(\vecx\ast K_x\big)^2+\big(\vecx\ast K_y\big)^2}.
\end{equation}

The outcome is thus a gray-scale image itself and the perturbation added to each color channel of the input image is identical.
Intuitively, the Sobel filter detects changes in the horizontal direction with $K_x$, in the vertical direction by $K_y$ and combines their magnitude. This simple filter is related to
the Prewitt filter~\cite{prewitt70} and the Scharr filter~\cite{scharr14}, which change the parameter values. Extensions to a filter sizes of $5\times 5$ and more
accurate approximations~\cite{KekGha10} also exist, however, for the sake of simplicity, we stick with the Sobel filter to evaluate our edge filter
attack for generating adversarial examples.

\subsection{Adversarial Filter Attack}
\label{sect:adaptive-filter-attack}
The previous attack is based on the idea that the edges are approximately equal to the gradient of the classification result with respect to the input. Since the employed edge filters are
based on a convolution operation, we can ask whether there exist better filters to craft adversarial examples. In other words, is there a filter that particularly impairs the performance of learning models in computer vision?

Given a convolutional filter $\mathbf{k}$ we can search for its optimal values by
formulating an optimization problem given by
\begin{equation}
\label{eq:adaptive-untargeted}
    \argmin_{\mathbf{k}} - \sum_{\vecx\in D} \ell\Big(\vecx+ \vecx\ast \mathbf{k},f_\theta(\vecx),\theta\Big) + \lambda \lVert \mathbf{k}\rVert_2^2.
\end{equation}

In order to solve this problem we require a dataset $D$ that is representative for the training data and by definition this problem aims to craft untargeted adversarial examples since
we only increase the loss on the original label $f_\theta(x)$. The $L^2$ norm of the filter regularizes the perturbation strength, the loss term steers the success rate of the attack
and the parameter $\lambda$ balances both properties. If the learning model and the loss function are differentiable with respect to $\vecx$ we can initialize the filters with random
values and apply optimization schemes like SGD to solve the problem above. An example for a resulting filter is shown in \cref{fig:filter-examples} (e) and
we will discuss its properties and relations to other filters in \cref{sect:filter-analysis}.

Notice that we refrain from using the magnitude of the convolution outcome as in \cref{eq:mia-sobel} in order to allow negative values in
$\mathbf{\delta}$ that may help to achieve the goal easier. It might look like a natural extension to compute the perturbation with multiple filters $\mathbf{k}_1,\dots,\mathbf{k}_M$,
i.e., $\delta=\sum_{i=1}^M \vecx\ast \mathbf{k}_i$, however, due to the linearity of the convolution operator, the same perturbation can be achieved with a single filter summing up
the values of all filter. It is also straight forward to minimize the loss towards a target class $y_t(\vecx)$ that can be chosen by the attacker such that targeted adversarial
examples can be generated by optimizing
\begin{equation*}
    \argmin_{\mathbf{k}} \sum_{\vecx\in D} \ell\Big(\vecx+ \vecx\ast \mathbf{k},y_t(\vecx),\theta\Big) + \lambda \lVert \mathbf{k}\rVert_2^2.
\end{equation*}

In consequence, we obtain two new attack strategies that significantly reduce the knowledge and capabilities of the adversary. The first strategy is readily applicable and does not require any knowledge of the model and the input sample. The second strategy requires white-box access to the model. Due to the few parameters of the filter, however, a considerably lower number of training examples is needed to achieve good results.

\section{Evaluation}
\label{sect:eval}
We evaluate both attack strategies on three neural networks, namely VGG-13~\cite{SimZis15}, ResNet-50~\cite{HeZhaRenSun16} and Inception-V3~\cite{SzeVanIof+16} that were
trained on the ImageNet dataset~\cite{imagenet}. In order to simulate an attack during test time, we use the ImageNet V2 dataset~\citep{RecRoeSch+19} that is composed of
\numprint{10000} images disjoint of the ImageNet dataset. We test our attack scheme on \numprint{5000} randomly chosen images thereof and use the remaining \numprint{5000}
to optimize the filter in the adaptive attack. To evaluate the
attack performance we use the attack success rate, i.e., the fraction of images for which $f_\theta(\vecx) \neq f_\theta(\vecx + \mathbf{\delta})$. 

As a quantification of the stealthiness of
$\mathbf{\delta}$ we employ the \emph{peak signal-to-noise ratio}, a measure of difference between $\vecx$ and $\vecx + \mathbf{\delta}$. In our case, the PSNR thus becomes a function
of $\mathbf{\delta}$ and is defined by
$$ \text{PSNR}(\mathbf{\delta}) = 10\cdot \log_{10}\Big(3wh\cdot\frac{I_{\text{max}}^{\text{ }2}}{\lVert\mathbf{\delta}\rVert_2^2}\Big),$$
\noindent where $I_{\text{max}}$ is the maximum value a pixel can have, i.e., \numprint{255} in our experiments. The PSNR is measured in dB and usually applied for evaluating image compression methods, thus
a low PSNR indicates higher information loss or a stronger perturbation in our case.

\begin{figure}[htbp]
    \input{figures/mia-sobel}
    \caption{Success rate of the edge filter attack for Sobel- and Laplacian-of-Gaussian (LoG) filter (left) and PSNR between original image and the outcome of the attack (right). The dashed line indicate an attack where amplified Gaussian noise
    is added to the image such that the PSNR (left) or success rate (right) is equal to the corresponding Sobel filter attack outcome.}
    \label{fig:eval-sobel-attack}
\end{figure}
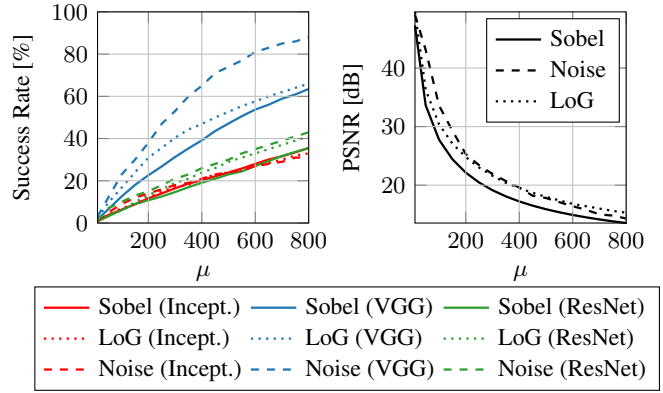

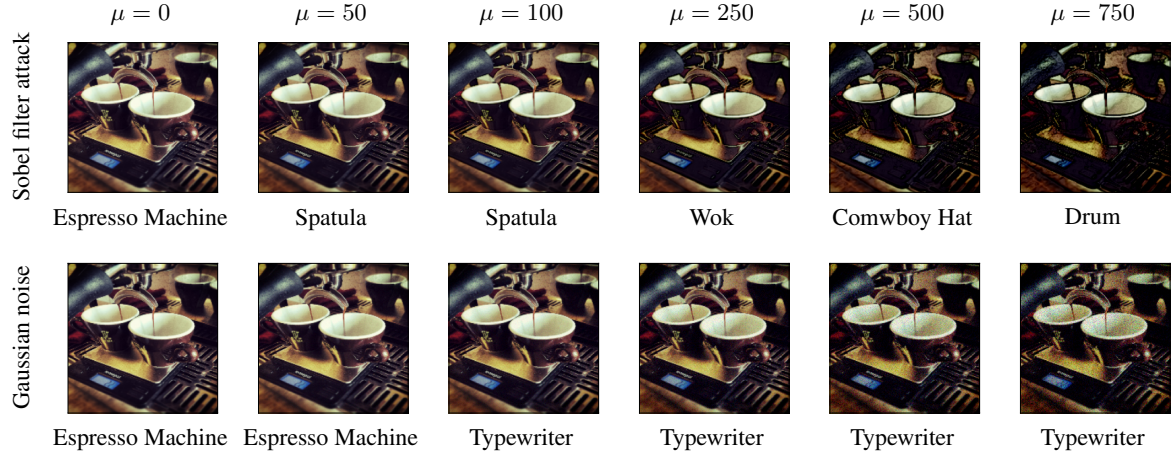
\begin{figure*}
    \centering
    \input{figures//trajectories}
    \caption{Outcome of the Sobel attack for different magnitudes $\mu$. The classification result is based on a VGG-13 model and is shown under the images. The Gaussian noise is computed
    such that the success rate is equal to the Sobel attack.}
    \label{fig:sobel-attack-trajectory}
\end{figure*}

\subsection{Edge Filter Attack}
\cref{fig:eval-sobel-attack} shows the evolution of the average attack success rate (left) and the PSNR (right) when increasing $\mu$ in the attack scheme above.
During our experiments we find that subtracting edge information ($\mu<0$) and adding it ($\mu>0$) achieves a similar performance with a slight advantage
for the subtraction, therefore we present only the results for $\mu<0$. As expected, the success rate of the attack rises at the cost of a lower PSNR as
$\mu$ is increased. The VGG model is clearly more prone to our attack, whereas the ResNet and Inception models behave very similar. To get a feeling for the
strength of the perturbations, we present a trajectory of the attack for the VGG model when increasing $\mu$ in \cref{fig:sobel-attack-trajectory}. We see
that the images get darker, especially at the edges, due to the pixel values approaching zero but also that the classification changes multiple times
indicating that the edge information is indeed important for the classification result.

As a baseline, we perform a simple attack where amplified Gaussian noise is subtracted from the input image, i.e.
$\mathbf{\delta} = -t\cdot\lvert\mathcal{N}(0,I)\rvert$ for some $t\in\mathbb{R}_+$. Here we use the absolute value of the noise since the edge filter attack
subtracts positive values due to the usage of the magnitude (see \cref{eq:sobel-filter}). Given a PSNR value of the Edge attack for some $\mu$,
we can perform a binary search over $t$ to find a value that achieves an equal PSNR value to compare the success rates at $\mu$. Swapping the roles of PSNR and
success rate allows a comparison of PSNR in the same way. 

The corresponding curves are presented with dashed lines in \cref{fig:eval-sobel-attack} and the resulting images are also depicted in
\cref{fig:sobel-attack-trajectory}. We see that the noise addition is better for the VGG model, equal for the ResNet model,
and slightly worse for the Inception model regarding both, success rate and PSNR. This result should, however, be taken with a grain of salt since the noise is applied uniformly and changes every single pixel in the image indiscriminately, whereas the edge attack is restricted to small fractions thereof, see the large white area in \cref{fig:intro-figure} for
example. Still, crafting adversarial examples by subtraction of edge information is possible but achieving high success rates is difficult. At the same
time, this simple attack required only 
$27$ parameters in total and still was able to fool all three networks with adversarial examples that were computed in a ''one shot'', training-free manner, requiring no iterative optimization.

\subsection{Adversarial Filter Attack}

The adversarial attack has multiple parameters that influence the result, e.g. size of the filters or the number of training points in $D$. To allow a close comparison to the
static attack investigated above and to be able to interpret the resulting filters, we use small filters of size three or five and optimize \cref{eq:adaptive-untargeted}
using SGD for $25$ epochs. We further split the \numprint{5000} images we have not been used so far into \numprint{3000} images that can be used for training and \numprint{2000} that 
will be used for validation. The \numprint{5000} examples used for the Sobel attack serve as our test set such that both attacks are compared on the same data.

The regularization strength $\lambda$ is the dominating parameter in the adversarial filter attack since it determines a success rate and a PSNR value that can be achieved during the
optimization process. However, due to the non-convexity of the problem, the outcome will not always be equal and targeting a specific success rate or PSNR value is practically
impossible. To this end, we optimize \cref{eq:adaptive-untargeted} for multiple values for $\lambda$. Concretely, we sample values between $\lambda=0$, i.e., high success rate and low
PSNR, and $\lambda=0.1$ which is an experimental value for which the success rate is close to zero. Sampling enough values allows to pick a given PSNR value for all
networks, for example, such that the success rates can be compared. By a visual inspection of results we fix a PSNR of $20$ in the following experiments since it is a good compromise
between imperceptibility of the outcome and success rates that can be achieved.

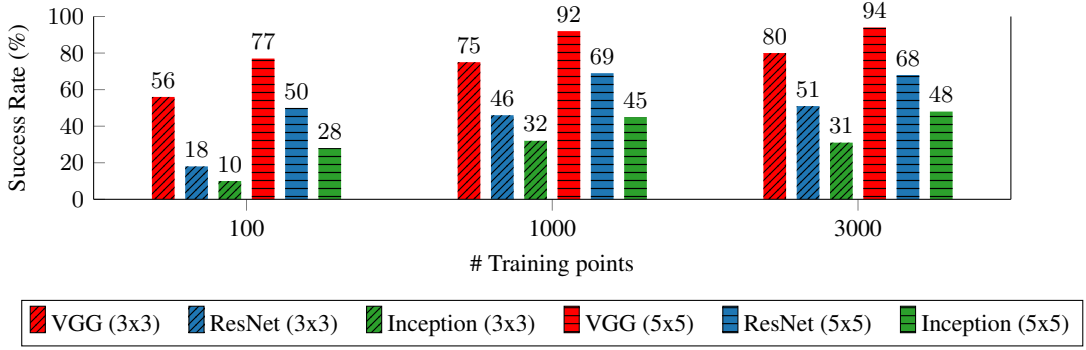
\begin{figure*}[h!]
    \centering
    \input{figures/mia-adaptive}
    \caption{Success rate of the adaptive filter attack for different numbers of training examples and filter sizes. The PSNR is fixed to $20$ for all models.}
    \label{fig:mia-adaptive}
\end{figure*}

\cref{fig:mia-adaptive} shows the success rates of the adversarial filter attack when varying the number of training examples between \numprint{100} and \numprint{3000} and for
convolution kernels of size $3\times 3$ and $5\times 5$. We observe that this attack achieves much better results compared to the simple Edge filter attack: For the VGG model, we can
increase the success rate from 27~\% to 56~\% with only \numprint{100} training examples. With \numprint{1000} training examples, the success rate of the ResNet- and Inception model
are increased by a factor of two and three respectively. Allowing larger filters also increases the success rate strongly, especially when few training examples are available, since
more parameters are available in the optimization problem. Likewise, more training examples also help given a fixed filter size although the effect becomes weaker when moving from
\numprint{1000} to \numprint{3000} training points. In summary, we can conclude that we can craft adversarial examples with adaptive filters in an extremely efficient way once a
suitable dataset for optimizing the filters is available. Moreover, compared to similar approaches~\citep[e.g.][]{BalFis18}, the number of parameters is reduced by up to \emph{six orders of
magnitude}.

\begin{figure*}[h!]
    \centering
    %\small{
    \input{figures/filters}
    %}
    \caption{Resulting convolution filters of the adversarial filter attack after optimization with \numprint{2000} training examples.}
    \label{fig:mia-filters}
\end{figure*}

\subsubsection{Adversarial Class distribution}

Since our attack is performed in an untargeted way, it is worth investigating the classes, to
which the adversarial versions of the images are assigned by the networks. \cref{tab:class-distribution} shows the five most common outcome classes of successful attacks for the ResNet
model for two PSNR values where the filters are optimized with \numprint{1000} examples. 

\begin{table}[h!]
\centering
\begin{tabular}{c|c|c|c}
    \hline
    \multicolumn{2}{c|}{$PSNR=20$} & \multicolumn{2}{c}{$PSNR=30$} \\ \cline{1-4}
    Class & Fraction & Class & Fraction \\ \hline
    swing           & 20.1\%       & window screen & 39.1\%       \\ \hline
    mountain bike   & 10.4\%       & strainer      & 1.6\%       \\ \hline
    dung beetle     & 3.0\%       & spider web    & 1.0\%       \\ \hline
    pinwheel        & 2.8\%       & harp          & 0.8\%       \\ \hline
    paddle          & 2.5\%       & shopping basket & 0.7\%     \\ %\hline
    % Add more rows as needed
\end{tabular}
\caption{Fraction of class outcomes of the adversarial examples for the ResNet with a filter
of size $5\times 5$.}
\label{tab:class-distribution}
\end{table}

Since the ImageNet dataset constitutes \numprint{1000} classes, a uniform distribution would yield
a fraction of $0.1\%$ for each class. However, we find that the distribution has a heavy tail
where the heaviness depends on the regularization strength or the PSNR value achieved. The total
number of classes appearing for successful attacks is X ($PSNR=20$) and Y ($PSNR=30$) indicating
that a large fraction of classes has not been targeted during optimization. Moreover, for a
weaker regularization (lower PSNR) the filter has more freedom resulting in about six target
classes that cover $40\%$ of the total appearing classes. For a stronger regularization (higher
PSNR), the task is more difficult and the outcome class is dominate by ``window screen`` which
covers $40\%$ alone. It is also interesting that the five top classes are disjoint for the
different regularization strengths, i.e. the dominating class for a $PSNR=30$ has been targeted
rarely for $PSNR=20$ and vice versa.

\subsubsection{Filter analysis}
\label{sect:filter-analysis}
As a first step towards understanding the optimized filters, we directly inspect their numerical values.  In \cref{fig:mia-filters} we present filters for
different color channels of size $3\times 3$ and $5\times 5$ from the previous experiment that were trained with \numprint{3000} examples. For a complete
comparison, all filters are shown in \cref{sect:apdpx-filters}. To compare filters across different models and sizes, we
normalize them to the range $[-1,1]$ by dividing all values by the largest absolute value.

By a first inspection we notice a lot of symmetries in the resulting filters. The $5\times 5$ filter has the highest value in the middle and symmetries along
the major axis. A specific derivation of its functionality, however, is difficult, especially due to the varying signings. The $3\times 3$ ResNet
filters are symmetric with respect to the middle row and column and the VGG filters are (approximately) symmetric with respect to the middle entry of the
matrix. While a concrete description of the functionality is difficult we can compare them to some well known kernels from image processing as presented in
\cref{fig:filter-examples}.
Firstly, since the middle row and middle column of the filters are not zero, the filters are different from the Sobel filters shown in \cref{eq:sobel-filter-raw}. A similar concept, however, can be observed
for the middle row of the VGG filters for channel one and three: The left and right neighbor pixels are subtracted from the current pixel with an (approximately) equal factor, resulting in an output of zero if the
pixel values are constant. Secondly, the filters are different from averaging- or Gaussian filters utilized for blurring. We notice,
however that the structure of the Gaussian kernel, i.e. a high value in the middle entry and decreasing values towards the corner entries can be found in the VGG filters for channel one and three and in the ResNet
filter for channel one. Taking the negative sign in the middle of the kernels into consideration, we can find relations to the Laplacian of a Gaussian: 
The corner values are negative wheres the border values are positive and (almost) twice the size of the corner values. The center value is also negative but
in the magnitude of the border values which sets the adversarial filters apart from the Laplacian of Gaussian.
%the ResNet model come very close to a Laplacian of Gaussian expect for the corner values which are a bit too large.

\subsubsection{Perturbation analysis}
\begin{figure*}[h!]
    \centering
    \input{figures/appendix-model-filters}
    \caption{Perturbations after convolving inputs with the adversarial filters optimized on different models.}
    \label{fig:mia-input-conv-filter}
\end{figure*}
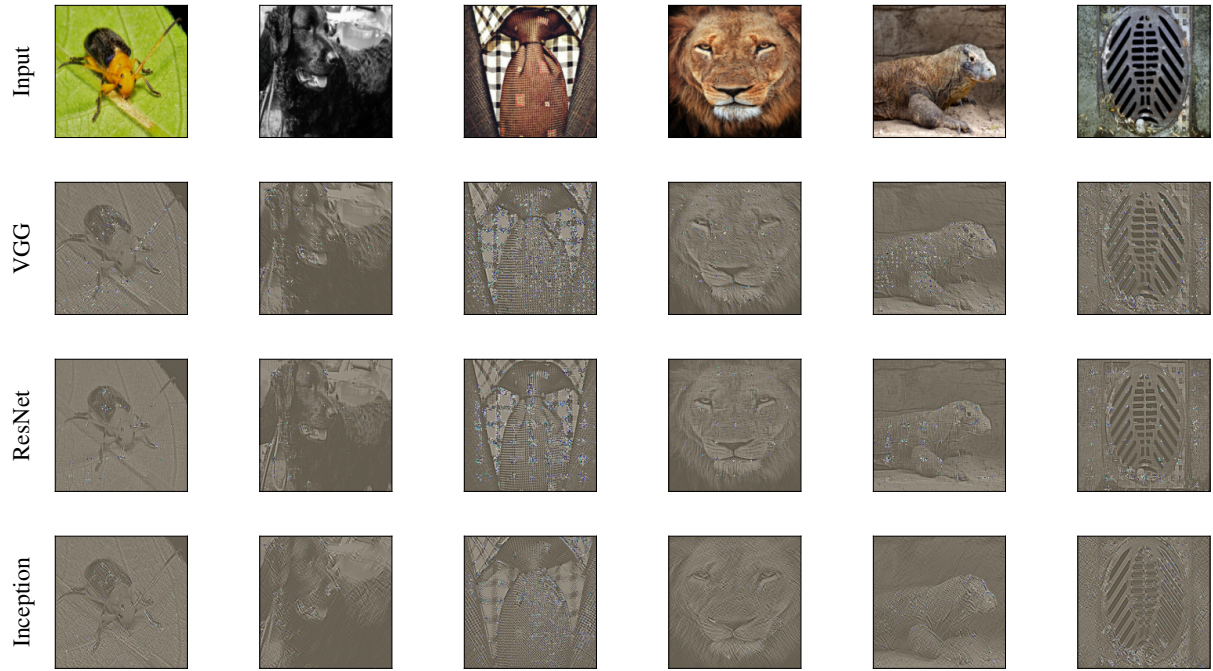

As a second step of evaluation, we investigate $\mathbf{\delta}$, i.e., the outcome of applying the optimized filters to the input image. The results are
presented in \cref{fig:mia-input-conv-filter} for the different model architectures from our experiments. We
observe that the outcome of the convolution is different to the classic adversarial perturbations that cover the entire image. Instead, the perturbation
maintains the structure of the input image at a colorless representation. Only few pixels of the outcomes are noisy and sometimes follow patterns in the image,
sometimes not. For example, the structure of the shirt in the third column is highlighted but all the structure in the last column is left. For the other
columns, the perturbations even seem randomly distributed over the image. Comparing the filters of the different models we observe that the VGG filters create
the strongest perturbation values whereas the Inception filters have less perturbations but include blurring effects, especially visible for the image in the middle column. Finding a pattern in the perturbations only by visual inspection is thus hard.

\begin{table*}[h!]
  \centering
  \begin{tabular}{c|ccc|ccc}
            \toprule
			\multirow[t]{2}{*}{\textbf{Filter training model}} & \multicolumn{3}{c|}{\textbf{PSNR=20}} & \multicolumn{3}{c}{\textbf{PSNR=30}} \\
			\cmidrule{2-7}
			& VGG & ResNet & Inception & VGG & ResNet & Inception\\
			\midrule
			VGG & 100\% & 102\% & 96\% & 100\% & 98\% & 102\% \\
			ResNet & 93\% & 100\% & 82\% & 50\% & 100\% & 51\%\\
			Inception & 86\% & 69\% & 100\% & 91\% & 75\% & 100\%\\
			\bottomrule
		\end{tabular}
  \caption{Relative attack success rate when using the adversarial filters for models they were not optimized for.}
  \label{tab:mia-transferability}
\end{table*}

\subsubsection{Filter Transferability}

Although our adversarial filter attack achieves better success rates, we currently assume that the adversary knows the model she is optimizing the filter for,
which is an unrealistic assumption in reality. For white-box adversarial perturbations, it is well known that
the adversarial property of perturbations is transferred over different model architectures~\citep[e.g.][]{SzeZarSut+14,CubZopScho+17,TraPapGoo+17}. In \cref{tab:mia-transferability} we report the success rate of the optimized
kernels when applying them to models they were not trained for. Specifically, we use kernels of size $3\times 3$ for all models, two levels of perturbation strength, i.e., $\text{PSNR}\in\{20,30\}$ and report relative success rates with respect to the optimal model, i.e., a filter optimized for a specific model
has $100\%$ success rate on it.

In general, we observe that the adversarial property of the filters transfers across the three models evaluated during our experiments. The filters for the ResNet- and VGG model perform very equal for
a PSNR value of $20$, which might be rooted in the similarity of their values discussed above. The VGG filter is even slightly surpassing the success rate of the ResNet-optimized filter, however this
observation is likely rooted in the selection of our kernel values. Recall that we sampled multiple values of $\lambda$ when solving the optimization problem in \cref{eq:adaptive-untargeted} and chose
PSNR values close to a target value from the corresponding models to compare them. The PSNR value of the VGG filter is slightly lower ($19.9$) compared to the ResNet one ($20.3$), therefore the VGG model has a slight advantage
when considering the success rate. Interestingly, the Inception filter which achieved the lowest success rate throughout the experiments is still highly effective for the VGG- and ResNet model and comes close
to the success rates of the optimal filters. In absolute values, the success rate for the ResNet model is almost three times higher than on the Inception model itself. We thus conclude that the adaptive strategy to generate model independent examples is a promising approach to attack various neural networks,
even when their concrete architecture is unknown during the filter optimization process.

\subsubsection{Adversarial example inspection}

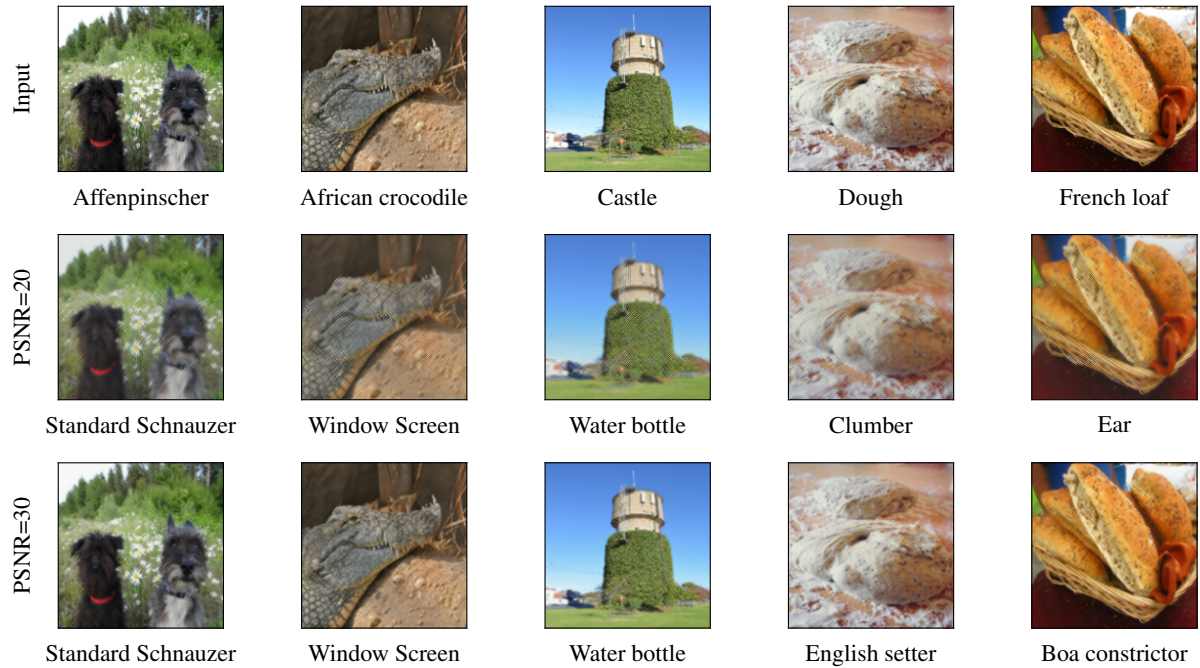
\begin{figure*}
\centering
    \input{figures/mia-adversarial-examples}
    \caption{Adversarial examples resulting from the adversarial filter attack on the ResNet. Classification is given below the images.}
    \label{fig:mia-adversarial-examples}
\end{figure*}

As a final evaluation step, we investigate the adversarial examples generated by the filters for two different perturbation strengths, i.e., $\text{PSNR}\in\{20,30\}$, visually as presented in \cref{fig:mia-adversarial-examples}.
Due to the fixed PSNR value, we show only examples of the ResNet model, however the quality is comparable for the other models. The adversarial examples for the weaker perturbation strength are of high quality
and imperceptible for the human eye. For a PSNR value of $20$, the perturbations are more visible and we can spot punctual blurring effects
in the adversarial examples.

Since our attack was performed in an untargeted way, it is worth investigating predictions for the adversarial examples. Firstly,
we find some pictures where the new class related to the original class, like in the leftmost column below where
one dog race is replaced by another one. However, we also find
predictions that are completely different from the original ones as in the other columns.

\section{Conclusion and Future Work}
Adversarial examples were a dominating research field in the security- and machine learning domain in the last decade. We found yet another way
to craft these imperceptible perturbations using the connection between explanations and classic edge detection strategies from computer vision.
Optimizing convolutional kernels we showed that perturbations achieving success rates of more than 90\% with a reasonable high PSNR value can be crafted
efficiently. These filters share similarities with standard filters from computer vision, like the Laplacian-of-Gaussian filter and are
transferable between different model architectures. Despite their symmetric structure, an explanation for their concrete workings and patterns to which they
respond is yet to be found. An interesting way to approach this problem would be the comparison to adversarial filters trained on different datasets, for
example with lower resolution, more structure in the images or fewer output classes.

Since there exist multiple approaches using architectures with millions of parameters to
craft adversarial perturbations in a similar way~\cite{BalFis18}, our results should be seen as a lower bound for the complexity of crafting model independent
adversarial examples. It seems likely that there is a ,,sweet spot'' not to far away from our approach that allows creating highly imperceptible
and highly effective adversarial examples with a fraction of parameters compared to the state of the art. A natural extension to our approach is thus the usage
of sequences of convolutions, maybe equipped with non-linear transformation functions, to slowly move our approach towards network architectures. The drastic
reduction in run-time to generate adversarial examples also may help to extend adversarial training~\cite{MadMakSch+18, ShaNajGhi+19} and thereby increase the
robustness of neural networks.

\bibliography{references}
\bibliographystyle{icml2022}

%%%%%%%%%%%%%%%%%%%%%%%%%%%%%%%%%%%%%%%%%%%%%%%%%%%%%%%%%%%%%%%%%%%%%%%%%%%%%%%
%%%%%%%%%%%%%%%%%%%%%%%%%%%%%%%%%%%%%%%%%%%%%%%%%%%%%%%%%%%%%%%%%%%%%%%%%%%%%%%
% APPENDIX
%%%%%%%%%%%%%%%%%%%%%%%%%%%%%%%%%%%%%%%%%%%%%%%%%%%%%%%%%%%%%%%%%%%%%%%%%%%%%%%
%%%%%%%%%%%%%%%%%%%%%%%%%%%%%%%%%%%%%%%%%%%%%%%%%%%%%%%%%%%%%%%%%%%%%%%%%%%%%%%
\appendix
\onecolumn
\section{Optimized filters for different models}
\label{sect:apdpx-filters}
In \cref{fig:filters-appendix} we show the optimized filters for the VGG and Inception model.

\begin{figure}[h!]
    \centering
    \input{figures/filters-appendix}
    \caption{Adversarial filters of size $3\times 3$ and a PSNR value of $20$ for all models.}
    \label{fig:filters-appendix}
\end{figure}

\begin{figure}[h!]
    \centering
    \input{figures/filters-appendix-5x5}
    \caption{Adversarial filters of size $5\times 5$ for a PSNR value of $20$ for the ResNet architecture.}
    \label{fig:filters-appendix-5}
\end{figure}

%%%%%%%%%%%%%%%%%%%%%%%%%%%%%%%%%%%%%%%%%%%%%%%%%%%%%%%%%%%%%%%%%%%%%%%%%%%%%%%
%%%%%%%%%%%%%%%%%%%%%%%%%%%%%%%%%%%%%%%%%%%%%%%%%%%%%%%%%%%%%%%%%%%%%%%%%%%%%%%

% This document was modified from the file originally made available by
% Pat Langley and Andrea Danyluk for ICML-2K. This version was created
% by Iain Murray in 2018, and modified by Alexandre Bouchard in
% 2019 and 2021 and by Csaba Szepesvari, Gang Niu and Sivan Sabato in 2022.
% Modified again in 2023 and 2024 by Sivan Sabato and Jonathan Scarlett.
% Previous contributors include Dan Roy, Lise Getoor and Tobias
% Scheffer, which was slightly modified from the 2010 version by
% Thorsten Joachims & Johannes Fuernkranz, slightly modified from the
% 2009 version by Kiri Wagstaff and Sam Roweis's 2008 version, which is
% slightly modified from Prasad Tadepalli's 2007 version which is a
% lightly changed version of the previous year's version by Andrew
% Moore, which was in turn edited from those of Kristian Kersting and
% Codrina Lauth. Alex Smola contributed to the algorithmic style files.

\end{document}

%% file: figures/intro/adv-example-eagle.tex
\begin{tikzpicture}[baseline,remember picture]

\newcommand{\signsize}{0.32\columnwidth}
\newcommand{\imdist}{0.01cm}
\newcommand{\rowdist}{0.1em}
\newcommand{\labeldist}{2.6cm}

\node (11) {\includegraphics[width=\signsize]{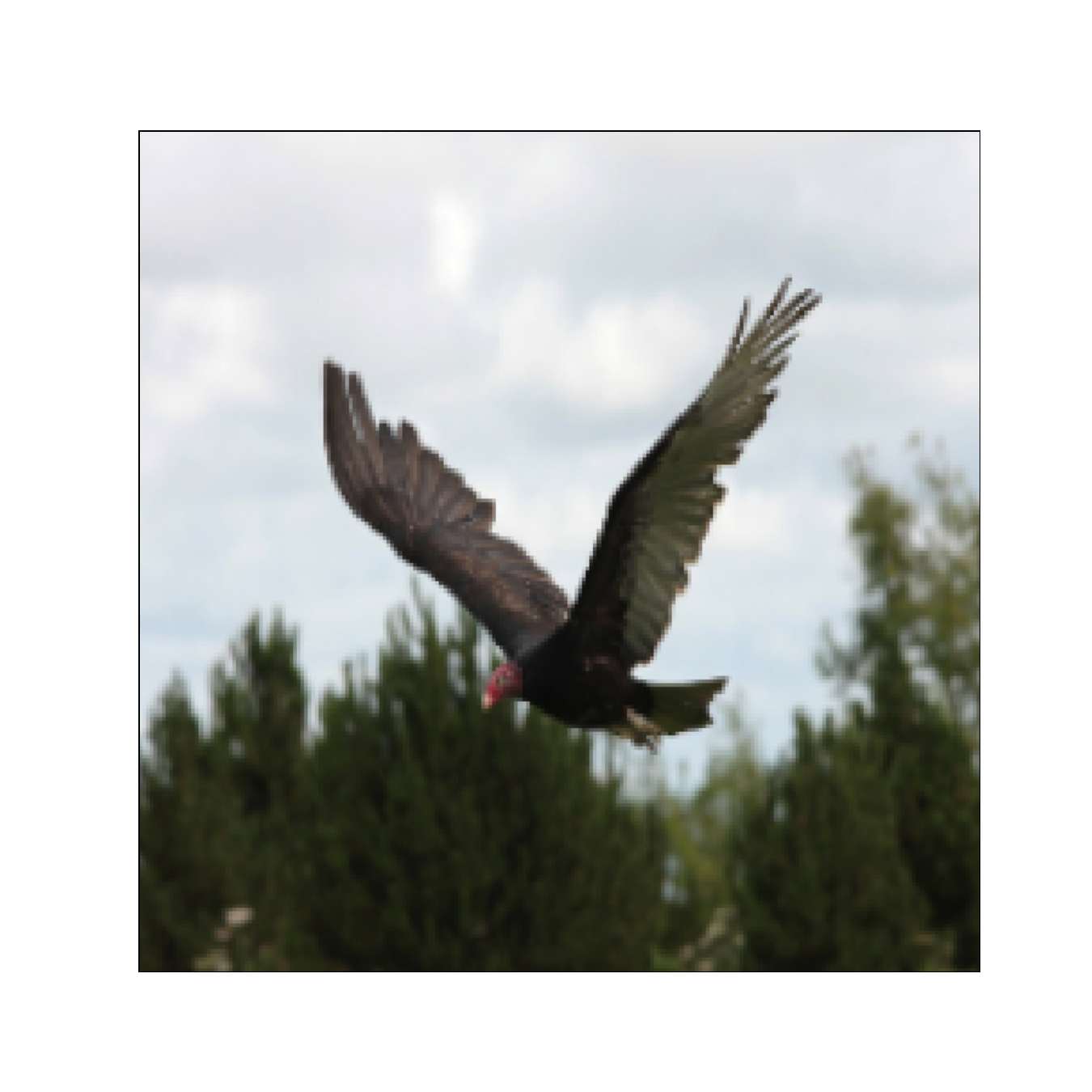}};
\node[below=\labeldist of 11.north] {\footnotesize{Prediction: Eagle}};
\node[right=\imdist of 11.east, xshift=-0.2cm](plus){\Large{-}};
%\node[right=0.4cm of plus.east](eps){\large $\epsilon\times$};
\node[right=\imdist of plus, xshift=-0.2cm] (12) {\includegraphics[width=\signsize]{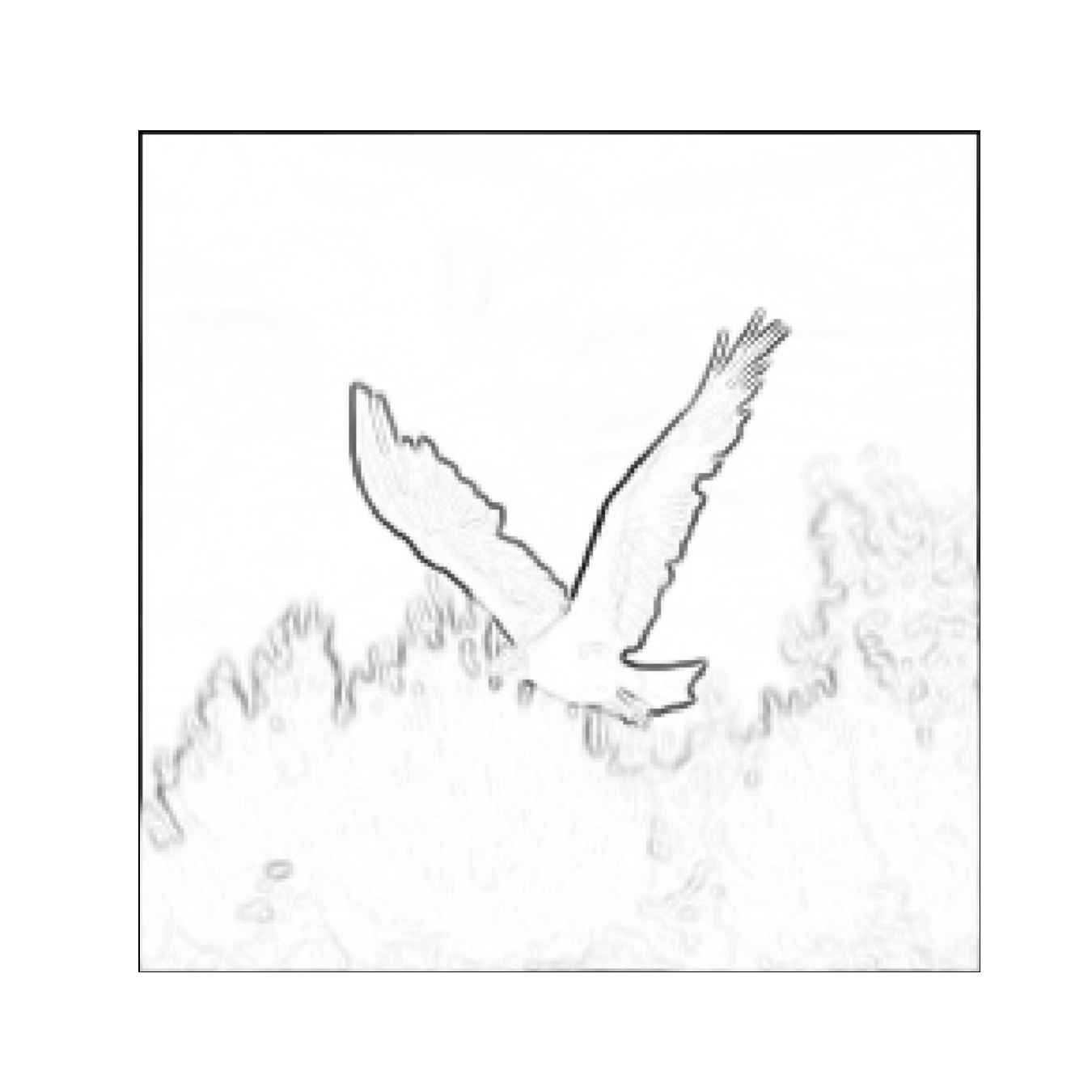}};
\node[below=\labeldist of 12.north] {\footnotesize{Edges of input}};
\node[right=\imdist of 12.east, xshift=-0.2cm](equal){\Large{=}};

%\node[right=\imdist of Speed] (Roundabout) {\includegraphics[width=\signsize]{data-figures/chapter-6/adversarial-filters/lion.pdf}};
\node[right=\imdist of equal, xshift=-0.2cm] (13) {\includegraphics[width=\signsize]{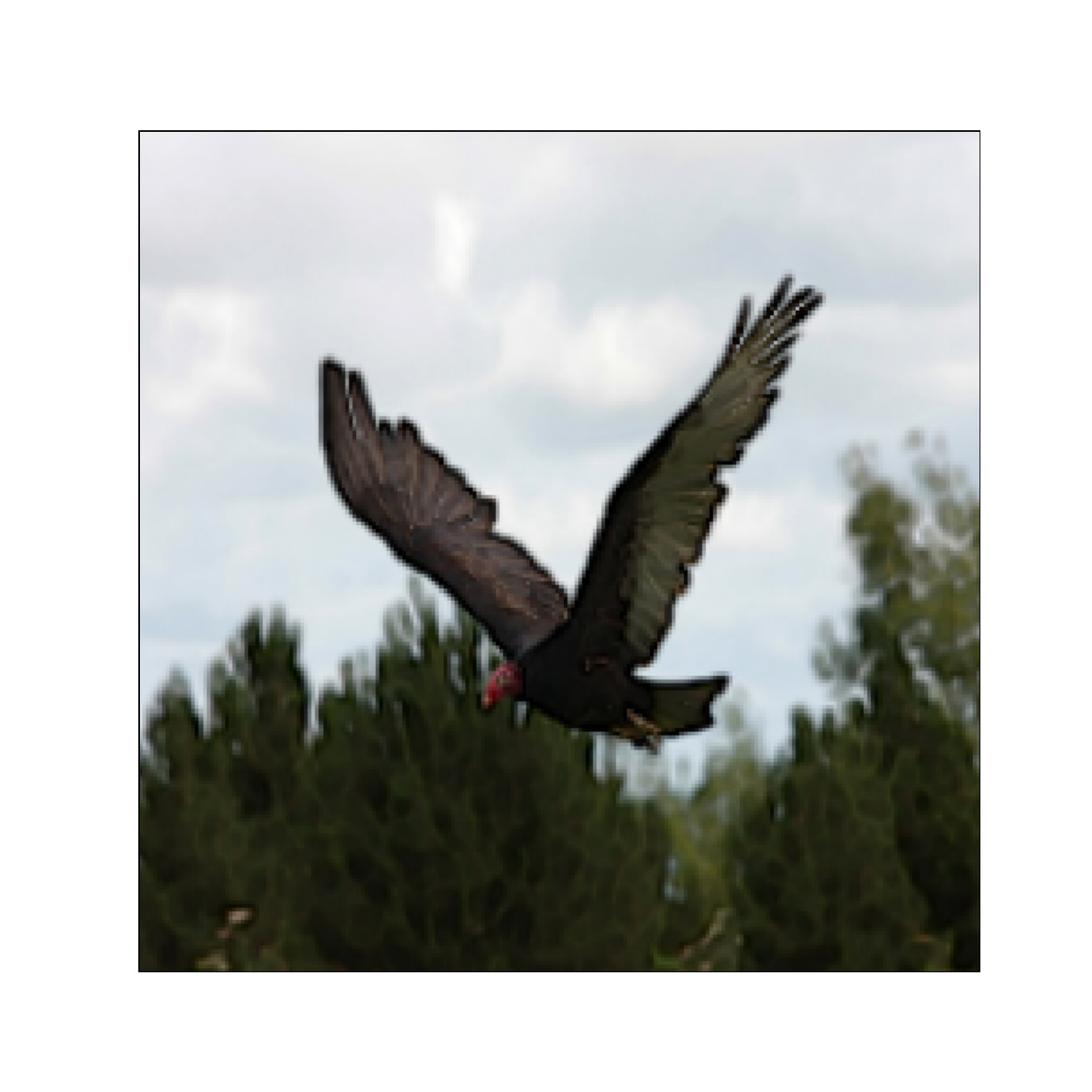}};
\node[below=\labeldist of 13.north] {\footnotesize{Prediction: Kite}};

\end{tikzpicture}

%% file: figures/intro-figure-grid/intro-mia.tex
\begin{tikzpicture}[baseline,remember picture]

\newcommand{\signsize}{0.12\textwidth}
\newcommand{\imdist}{0.1cm}
\newcommand{\rowdist}{0.1em}
\newcommand{\labeldist}{2.2cm}

\node (11) {\includegraphics[width=\signsize]{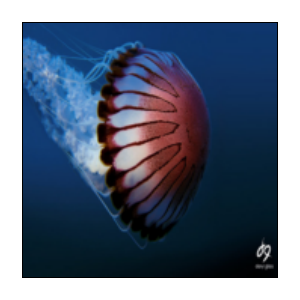}};
\node[left=0.4cm of 11.west, anchor=north, rotate=90]{\footnotesize{Input}};
\node[right=\imdist of 11] (12) {\includegraphics[width=\signsize]{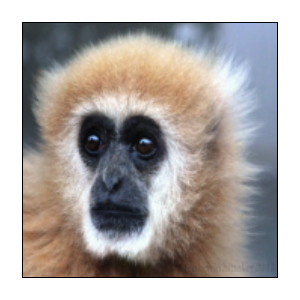}};
%\node[right=\imdist of Speed] (Roundabout) {\includegraphics[width=\signsize]{figures/intro-figure-grid/Crocodile.pdf}};
\node[right=\imdist of 12] (13) {\includegraphics[width=\signsize]{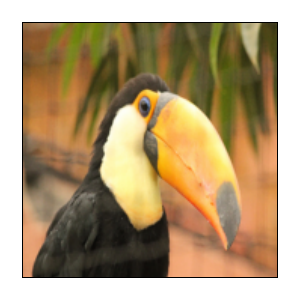}};
%\node[right=\imdist of ROW] (Backdoor) {\includegraphics[width=\signsize]{figures/intro-figure-grid/Cham.pdf}};

\node[below=\rowdist of 11.south] (21) {\includegraphics[width=\signsize]{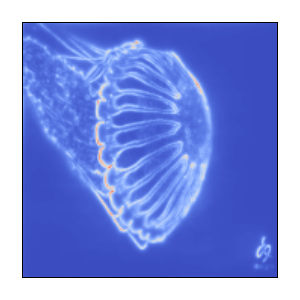}};
\node[left=0.4cm of 21.west, anchor=north, rotate=90]{\footnotesize{Explanation}};
\node[right=\imdist of 21] (22) {\includegraphics[width=\signsize]{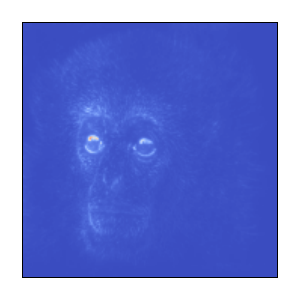}};
%\node[right=\imdist of Speed-ex] (Roundabout-ex) {\includegraphics[width=\signsize]{figures/intro-figure-grid/Crocodile-ex.pdf}};
\node[right=\imdist of 22] (23) {\includegraphics[width=\signsize]{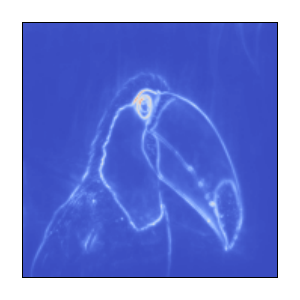}};
%\node[right=\imdist of ROW-ex] (Backdoor-ex) {\includegraphics[width=\signsize]{figures/intro-figure-grid/Cham-ex.pdf}};

\node[below=\rowdist of 21.south] (31) {\includegraphics[width=\signsize]{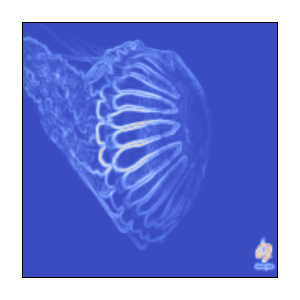}};
\node[left=0.4cm of 31.west, anchor=north, rotate=90]{\footnotesize{Edges}};
\node[right=\imdist of 31] (32) {\includegraphics[width=\signsize]{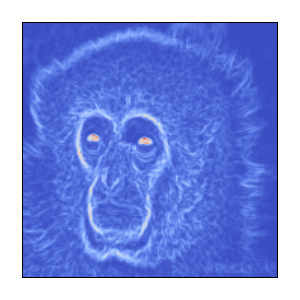}};
%\node[right=\imdist of Speed-edge] (Roundabout-edge) {\includegraphics[width=\signsize]{figures/intro-figure-grid/edges-Crocodile.pdf}};
\node[right=\imdist of 32] (33) {\includegraphics[width=\signsize]{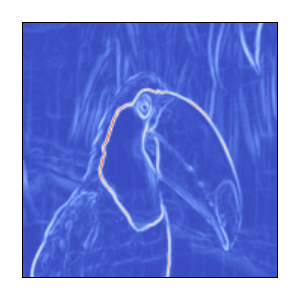}};
%\node[right=\imdist of ROW-edge] (Backdoor-edge) {\includegraphics[width=\signsize]{figures/intro-figure-grid/edges-Cham.pdf}};

\end{tikzpicture}

%% file: figures/blurring-filter.tex
\begin{tikzpicture}

\node(11) {$\frac{1}{9}\begin{bmatrix}
		1&1&1\\
        1&1&1\\
        1&1&1
	\end{bmatrix}$};

\node[above=of 11, yshift=-0.8cm] {\textbf{Smoothing}};
\node[below=of 11, yshift=0.9cm, xshift=0.1cm] {(a)};

\node[right=of 11, xshift=-0.1cm] (12) {$\frac{1}{16}\begin{bmatrix}
		1 & 2 & 1\\
        2 & 4 & 2\\
        1 & 2 & 1
	\end{bmatrix}$};

 \node[above=of 12, yshift=-0.7cm] {\textbf{Gaussian}};
 \node[below=of 12, yshift=0.9cm, xshift=0.2cm] {(b)};

 \node[right=of 12, xshift=-0.1cm] (13) {$\begin{bmatrix}
		0 & 1 & 0\\
        1 & -4 & 1\\
        0 & 1 & 0
	\end{bmatrix}$};

 \node[above=of 13, yshift=-0.8cm] {\textbf{Laplacian}};
 \node[below=of 13, yshift=0.9cm] {(c)};

 \node[right=of 13, xshift=-0.1cm] (14) {$\begin{bmatrix}
		-1 & 2 & -1\\
        2 & -4 & 2\\
        -1 & 2 & -1
	\end{bmatrix}$};

 \node[above=of 14, yshift=-0.8cm] {\textbf{Laplacian of Gaussian}};
 \node[below=of 14, yshift=0.9cm] {(d)};

 \node[right=of 14, xshift=-0.1cm] (15) {$\begin{bmatrix}
		-0.54 & 0.80 & -0.53\\
		0.80 & -1.00 & 0.80\\
		-0.51 & 0.81 & -0.53
    \end{bmatrix}$};
\node[above=of 15, yshift=-0.8cm] {\textbf{Adversarial Filter (ours)}};
\node[below=of 15, yshift=0.9cm] {(e)};
 
\end{tikzpicture}

%% file: figures/mia-sobel.tex
\definecolor{blue}{HTML}{1f77b4}
\definecolor{orange}{HTML}{ff7f0e}
\definecolor{green}{HTML}{2ca02c}

\begin{tikzpicture}

\hspace{-.3cm}
\begin{groupplot}[group style = {group size = 2 by 1,
                                 horizontal sep = 40pt,
                                 vertical sep=40pt},
                                 xlabel={$\mu$},
                                 xmax=800,
                                 width = 0.53\columnwidth,
                                 height = 0.53\columnwidth,
                                 legend cell align={left},
                                 legend style = {column sep = 1pt, legend columns = 3, font=\footnotesize, at={(0.4\linewidth,-0.3)}, anchor=north},
                                 grid style={on layer=axis background},
                ]

    \nextgroupplot[
      title = {},
      ylabel style={align=center,font=\footnotesize},
      xlabel style={align=center,font=\footnotesize},
      tick label style={font=\footnotesize},
      ylabel={Success Rate [\%]},
      axis on top,
      ymin=0, ymax=100,
      %ytick={25,50,75,100},
      %tick label style={font=\small},
      %legend cell align={left},
      %legend pos=north west,
      enlarge x limits=0,
      enlarge y limits=0,
      grid = both
    ]
    \addplot[color=red, line width=0.3mm] plot [] table [x=tau, y expr=\thisrowno{1}*100, col sep=comma]{figures/Inception_sobel.csv};
    \addlegendentry{Sobel (Incept.)}

    \addplot[color=blue, line width=0.3mm] plot [] table [x=tau, y expr=\thisrowno{1}*100, col sep=comma]{figures/VGG_sobel.csv};
    \addlegendentry{Sobel (VGG)}

    \addplot[color=green, line width=0.3mm] plot [] table [x=tau, y expr=\thisrowno{1}*100, col sep=comma]{figures/ResNet_sobel.csv};
    \addlegendentry{Sobel (ResNet)}

    \addplot[color=red, line width=0.3mm,dotted] plot [] table [x=tau, y expr=\thisrowno{1}*100, col sep=comma]{figures/inception-laplace.csv};
    \addlegendentry{LoG (Incept.)}

    \addplot[color=blue, line width=0.3mm, dotted] plot [] table [x=tau, y expr=\thisrowno{1}*100, col sep=comma]{figures/VGG-laplace.csv};
    \addlegendentry{LoG (VGG)}

    \addplot[color=green, line width=0.3mm, dotted] plot [] table [x=tau, y expr=\thisrowno{1}*100, col sep=comma]{figures/ResNet-laplace.csv};
    \addlegendentry{LoG (ResNet)}

    \addplot[color=red, line width=0.3mm,dashed] plot [] table [x=tau, y expr=\thisrowno{6}*100, col sep=comma]{figures/Inception_sobel.csv};
    \addlegendentry{Noise (Incept.)}

    \addplot[color=blue, line width=0.3mm, dashed] plot [] table [x=tau, y expr=\thisrowno{6}*100, col sep=comma]{figures/VGG_sobel.csv};
    \addlegendentry{Noise (VGG)}

    \addplot[color=green, line width=0.3mm, dashed] plot [] table [x=tau, y expr=\thisrowno{6}*100, col sep=comma]{figures/ResNet_sobel.csv};
    \addlegendentry{Noise (ResNet)}

    %\addplot[color=black, line width=0.3mm] plot [] coordinates {(200,200)(400,400)};
    %\addlegendentry{Sobel}

    %\addplot[color=red, line width=0.3mm, dashed] plot [] table [x=tau, y expr=\thisrowno{2}*100, col sep=comma]{figures/Inception_sobel.csv};
    %\addlegendentry{Inception-V3 (channel-wise)}

    %\addplot[color=blue, line width=0.3mm, dashed] plot [] table [x=tau, y expr=\thisrowno{2}*100, col sep=comma]{figures/VGG_sobel.csv};
    %\addlegendentry{VGG-19 (channel-wise)}

    %\addplot[color=green, line width=0.3mm, dashed] plot [] table [x=tau, y expr=\thisrowno{2}*100, col sep=comma]{figures/ResNet_sobel.csv};
    %\addlegendentry{ResNet-50 (channel-wise)}

    \nextgroupplot[
      title = {},
      legend style = {legend columns = 1, font=\footnotesize},
      ylabel style={align=center,font=\footnotesize},
      xlabel style={align=center,font=\footnotesize},
      tick label style={font=\footnotesize},
      ylabel={PSNR [dB]},
      axis on top,
      %label style={font=\footnotesize},
      %ymin=0, ymax=1.5,
      %ytick={1,10,100, 1000},
      %tick label style={font=\small},
      %legend cell align={left},
      legend pos=north east,
      enlarge x limits=0,
      enlarge y limits=0,
      grid = both
    ]

    \addplot[color=black, line width=0.3mm] plot [] table [x=tau, y=psnr-classic, col sep=comma]{figures/Inception_sobel.csv};
    \addlegendentry{Sobel}

    %\addplot[color=blue, line width=0.3mm, dashed] plot [] table [x=tau, y=psnr-noise, col sep=comma]{figures/VGG_sobel.csv};
    %\addlegendentry{Noise (VGG)}

    \addplot[color=black, line width=0.3mm, dashed] plot [] table [x=tau, y=psnr-noise, col sep=comma]{figures/ResNet_sobel.csv};
    \addlegendentry{Noise}

    \addplot[color=black, line width=0.3mm, dotted] plot [] table [x=tau, y=psnr-classic, col sep=comma]{figures/resNet-laplace.csv};
    \addlegendentry{LoG}

    %\addplot[color=red, line width=0.3mm, dashed] plot [] table [x=tau, y=psnr-noise, col sep=comma]{figures/Inception_sobel.csv};
    %\addlegendentry{Noise (Inception)}
    %\addplot[color=black, line width=0.3mm, dashed] plot [] table [x=tau, y=psnr-cwise, col sep=comma]{figures/Inception_sobel.csv};

    %\addplot[color=blue, line width=0.3mm] plot [] table [x=tau, y=psnr-classic, col sep=comma]{figures/VGG_sobel.csv};

    %\addplot[color=green, line width=0.3mm] plot [] table [x=tau, y=psnr-classic, col sep=comma]{figures/ResNet_sobel.csv};

\end{groupplot}
%\node at ($(group c2r1) - (4.0cm ,-2.5cm)$) {\ref{grouplegend}}; 
\end{tikzpicture}

%% file: figures/trajectories.tex
\begin{tikzpicture}[baseline,remember picture]

\newcommand{\signsize}{0.15\textwidth}
\newcommand{\imdist}{1mm}
\newcommand{\rowdist}{1mm}
\newcommand{\labeldist}{2.5cm}
\newcommand{\xshift}{-0.4cm}

\node (im1) {\includegraphics[width=\signsize]{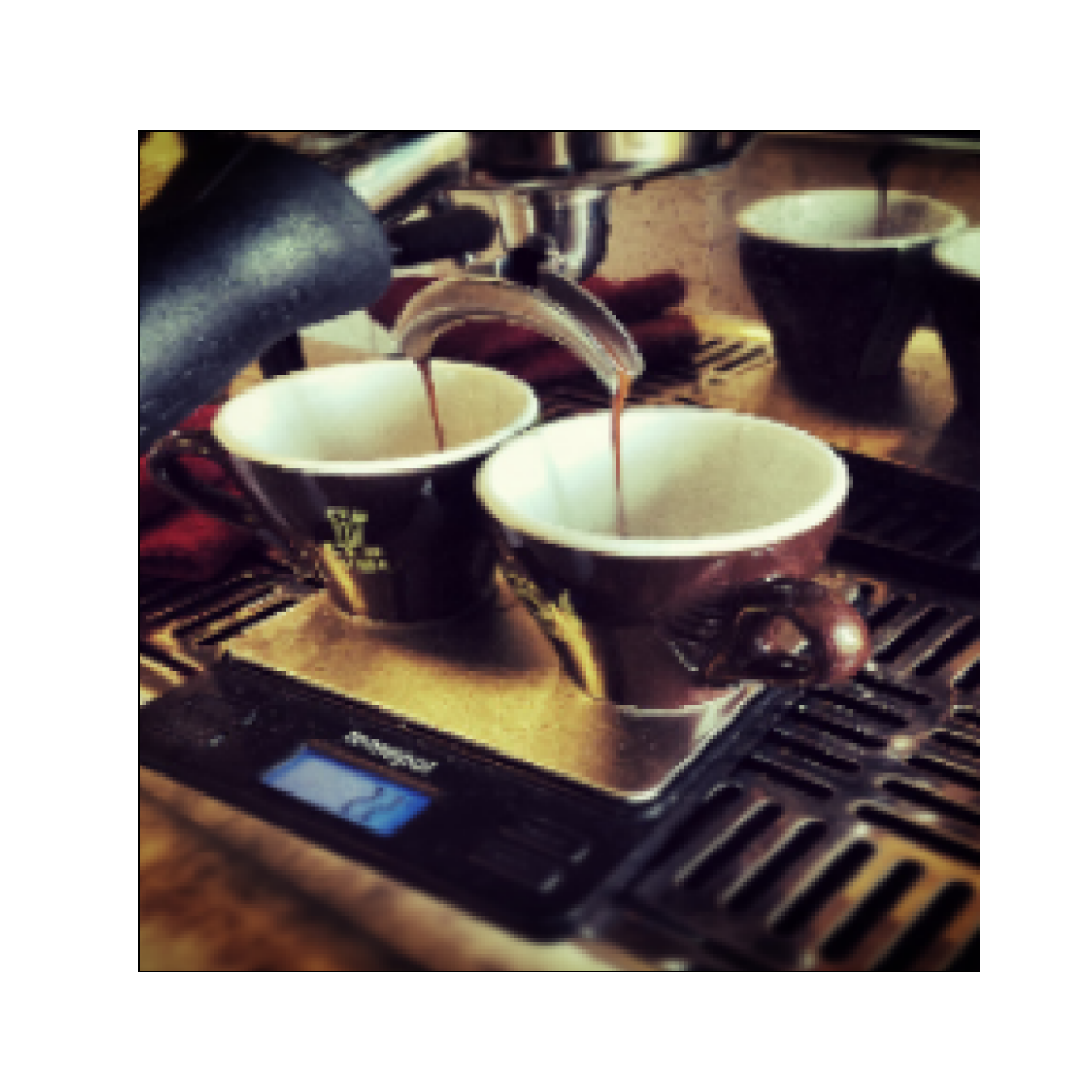}};
\node[left=0.4cm of im1.west, anchor=north, rotate=90]{\footnotesize{Sobel filter attack}};
\node[below=\labeldist of im1.north] {\footnotesize{Espresso Machine}};
\node[above=\labeldist of im1.south] {\footnotesize{$\mu=0$}};
\node[right=\imdist of im1, xshift=\xshift] (im2) {\includegraphics[width=\signsize]{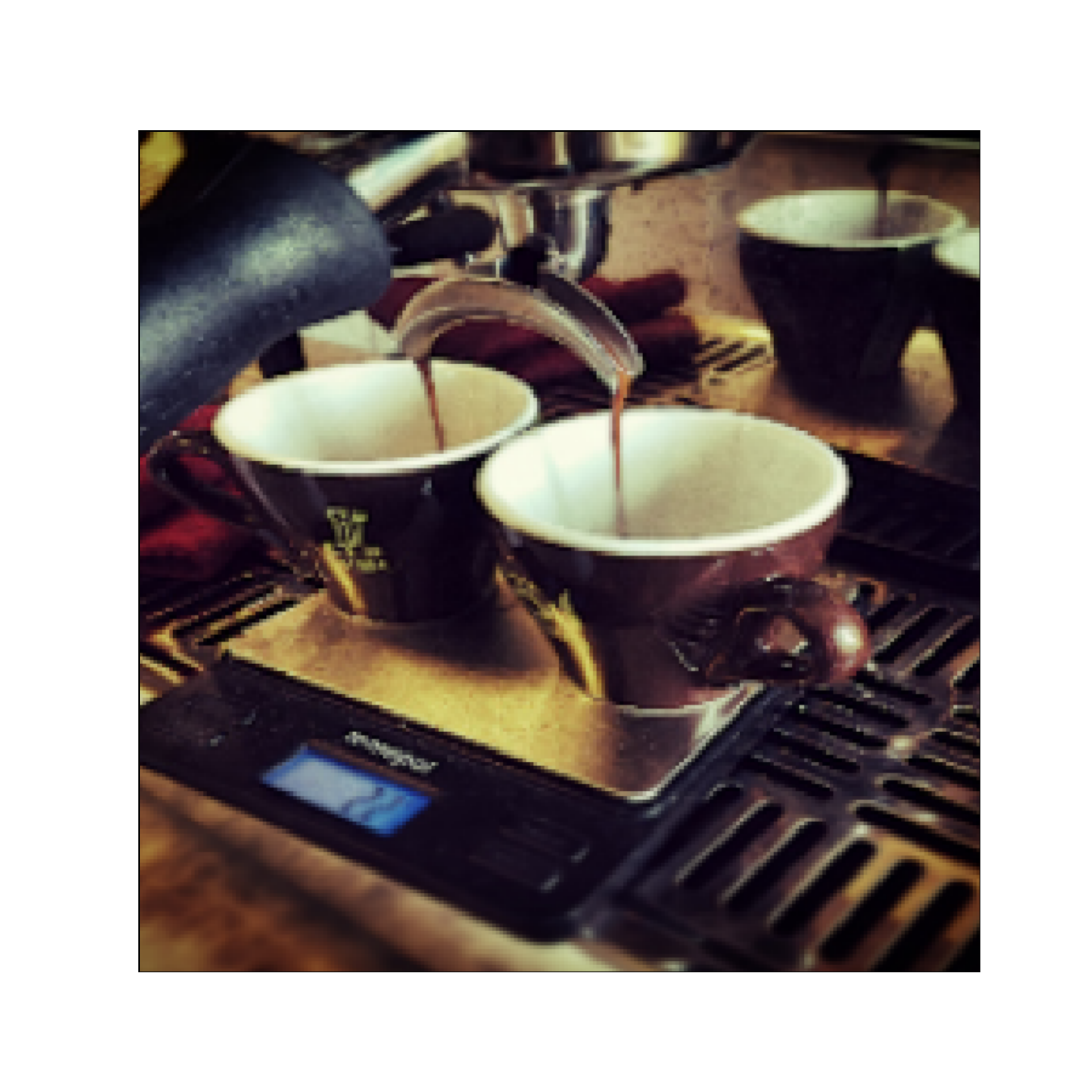}};
\node[below=\labeldist of im2.north] {\footnotesize{Spatula}};
\node[above=\labeldist of im2.south] {\footnotesize{$\mu=50$}};
\node[right=\imdist of im2, xshift=\xshift] (im3) {\includegraphics[width=\signsize]{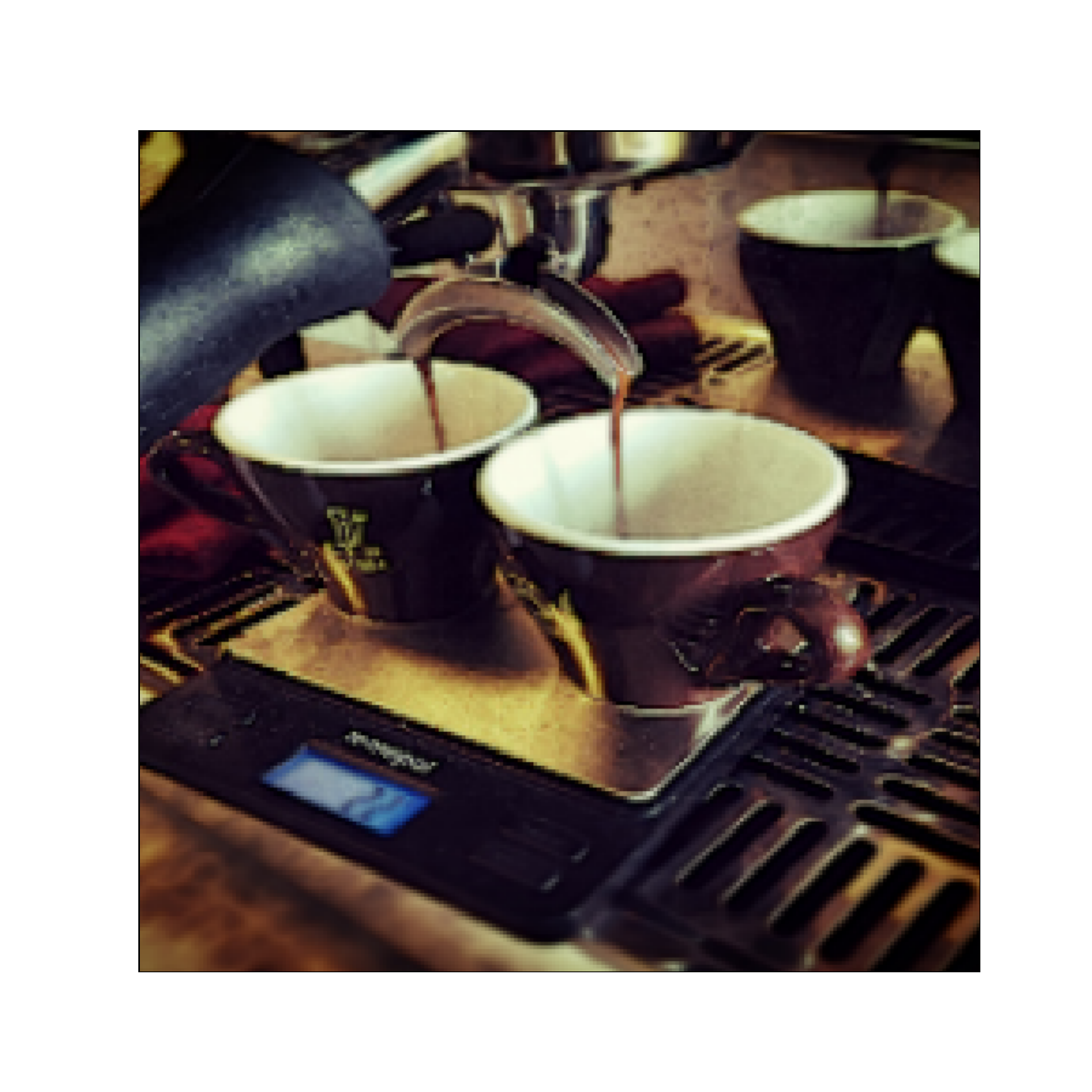}};
\node[below=\labeldist of im3.north] {\footnotesize{Spatula}};
\node[above=\labeldist of im3.south] {\footnotesize{$\mu=100$}};
\node[right=\imdist of im3, xshift=\xshift] (im4) {\includegraphics[width=\signsize]{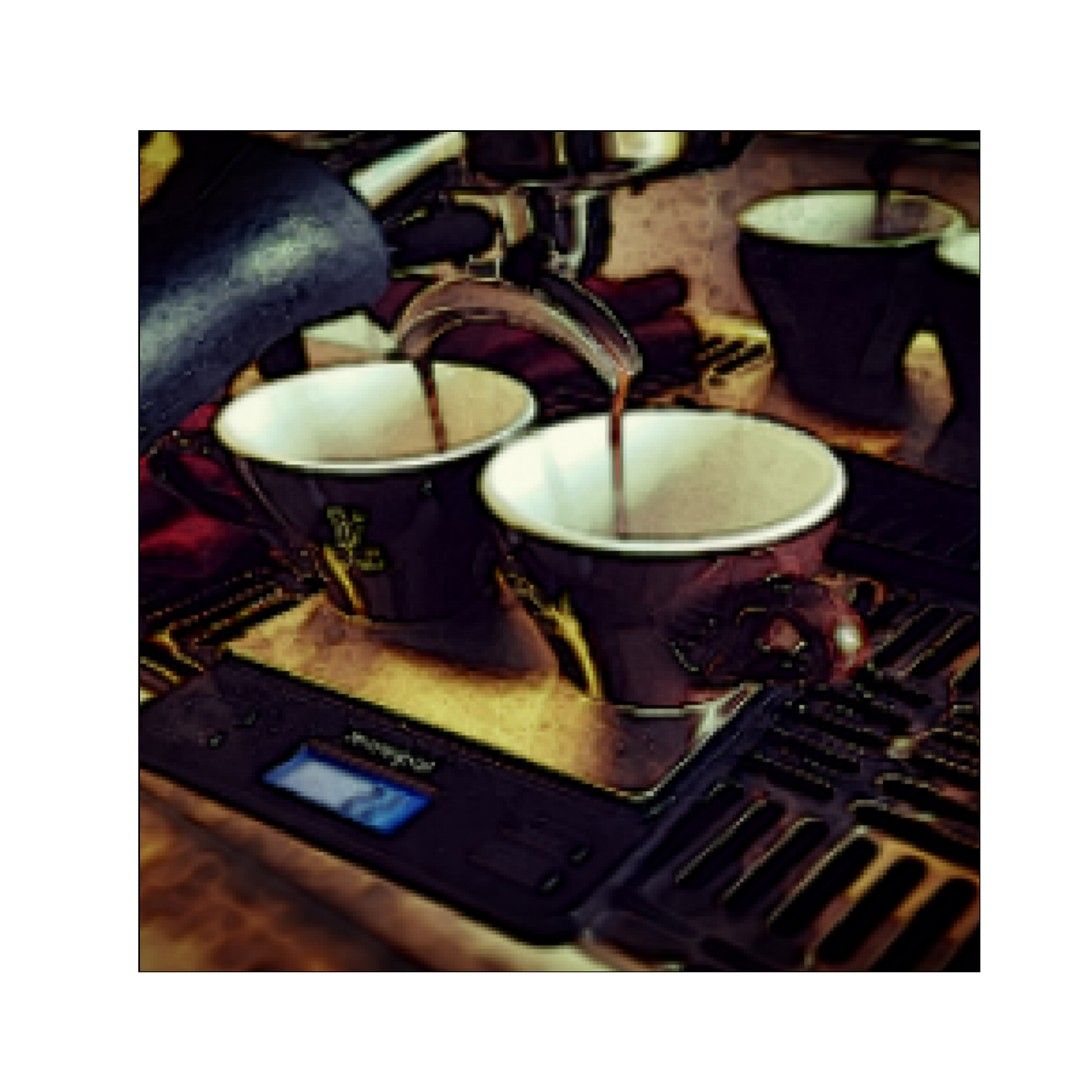}};
\node[below=\labeldist of im4.north] {\footnotesize{Wok}};
\node[above=\labeldist of im4.south] {\footnotesize{$\mu=250$}};
\node[right=\imdist of im4, xshift=\xshift] (im5) {\includegraphics[width=\signsize]{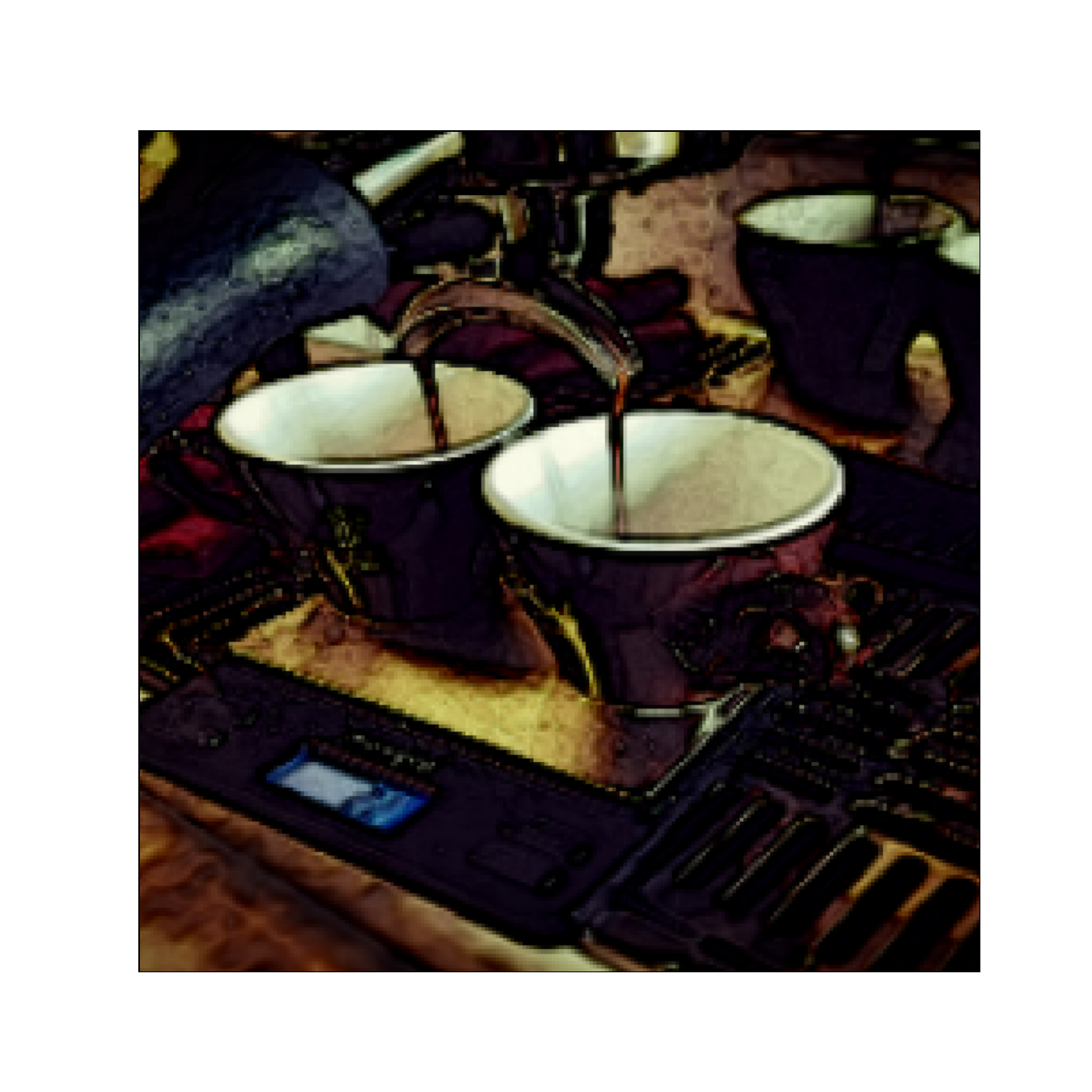}};
\node[below=\labeldist of im5.north] {\footnotesize{Comwboy Hat}};
\node[above=\labeldist of im5.south] {\footnotesize{$\mu=500$}};
\node[right=\imdist of im5, xshift=\xshift] (im6) {\includegraphics[width=\signsize]{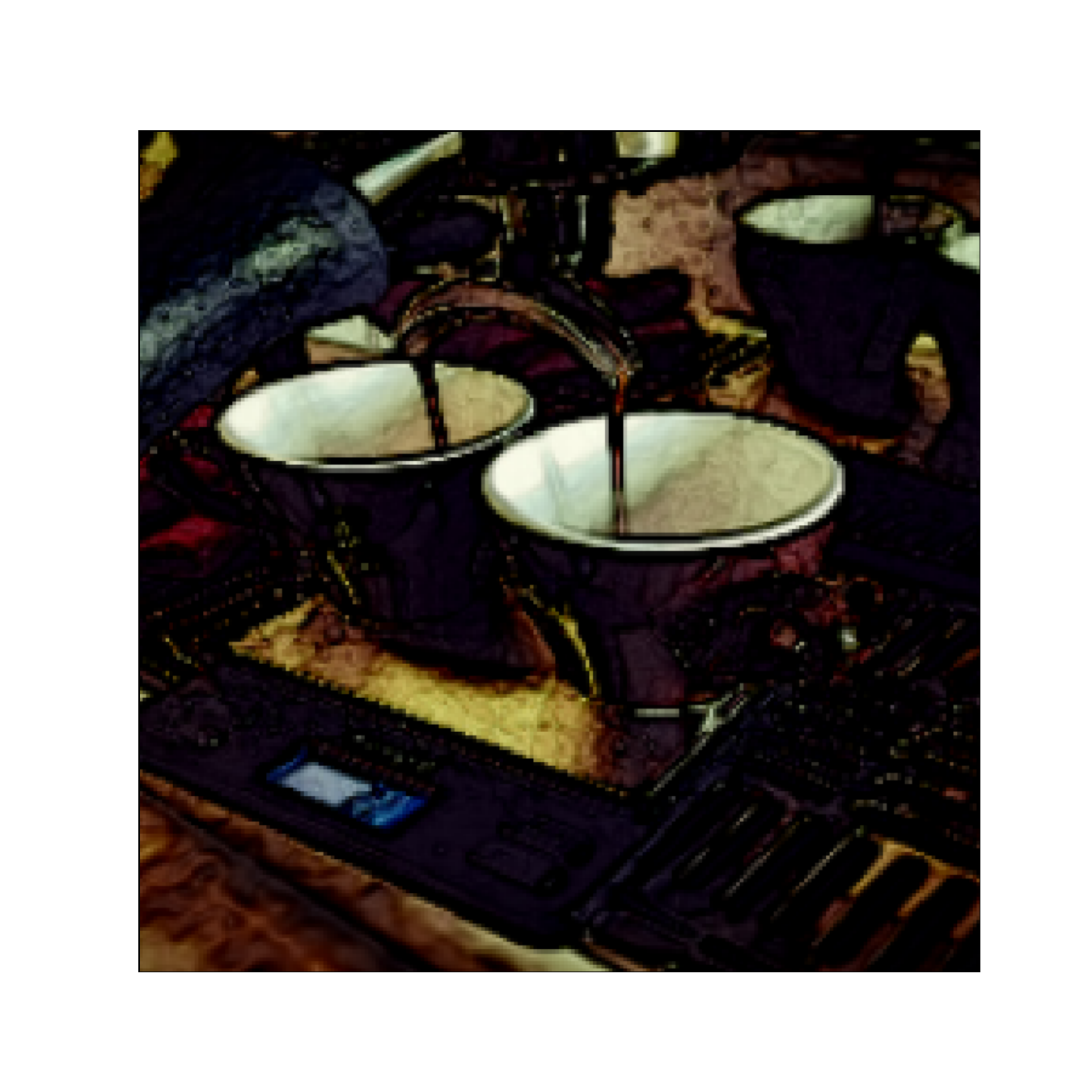}};
\node[below=\labeldist of im6.north] {\footnotesize{Drum}};
\node[above=\labeldist of im6.south] {\footnotesize{$\mu=750$}};

\node[below=\rowdist of im1.south] (row21) {\includegraphics[width=\signsize]{figures/trajectory/espresso/trajectory_0-adv-espresso_maker.pdf}};
\node[left=0.4cm of row21.west, anchor=north, rotate=90]{\footnotesize{Gaussian noise}};
\node[below=\labeldist of row21.north] {\footnotesize{Espresso Machine}};
\node[right=\imdist of row21, xshift=\xshift] (row22) {\includegraphics[width=\signsize]{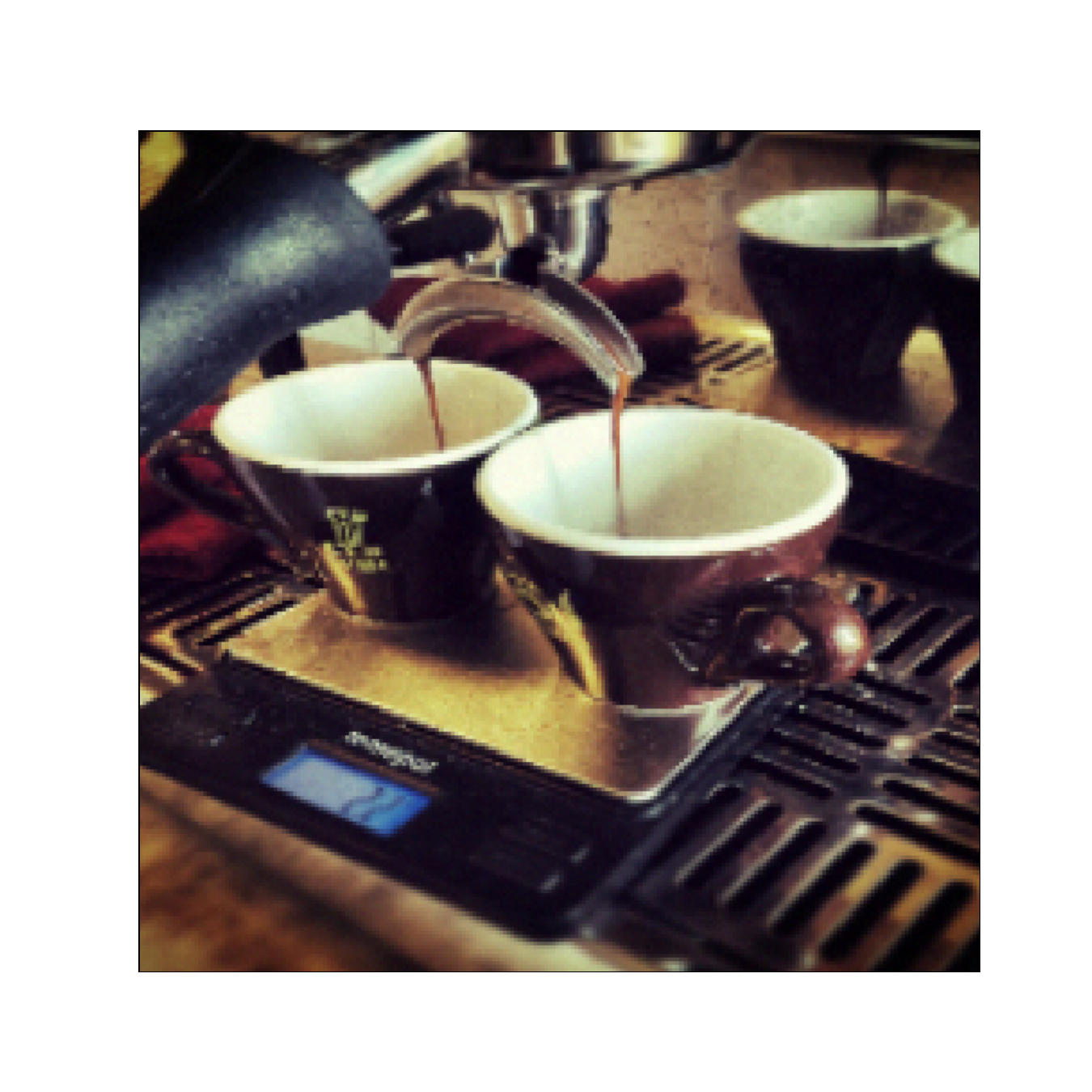}};
\node[below=\labeldist of row22.north] {\footnotesize{Espresso Machine}};
\node[right=\imdist of row22, xshift=\xshift] (row23) {\includegraphics[width=\signsize]{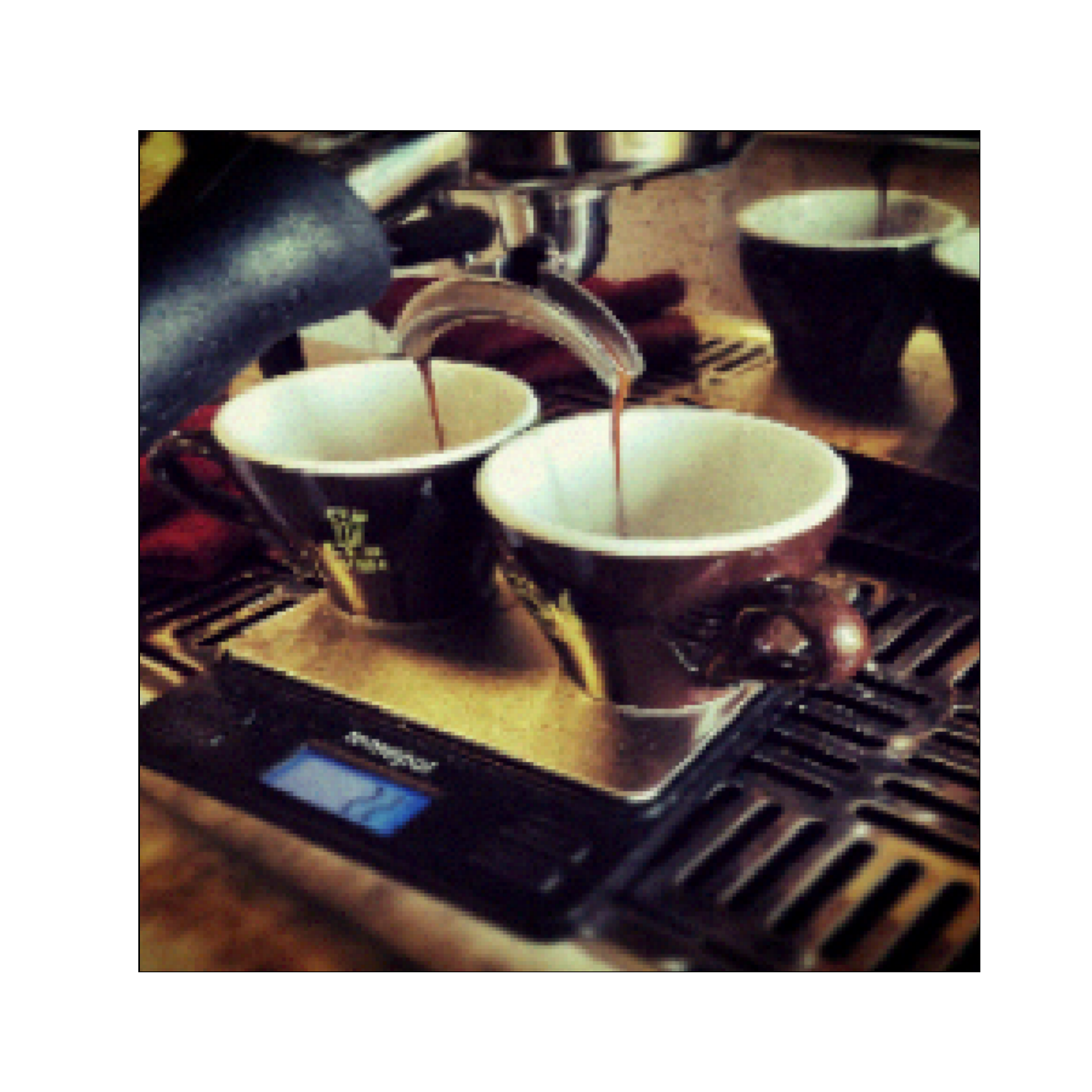}};
\node[below=\labeldist of row23.north] {\footnotesize{Typewriter}};
\node[right=\imdist of row23, xshift=\xshift] (row24) {\includegraphics[width=\signsize]{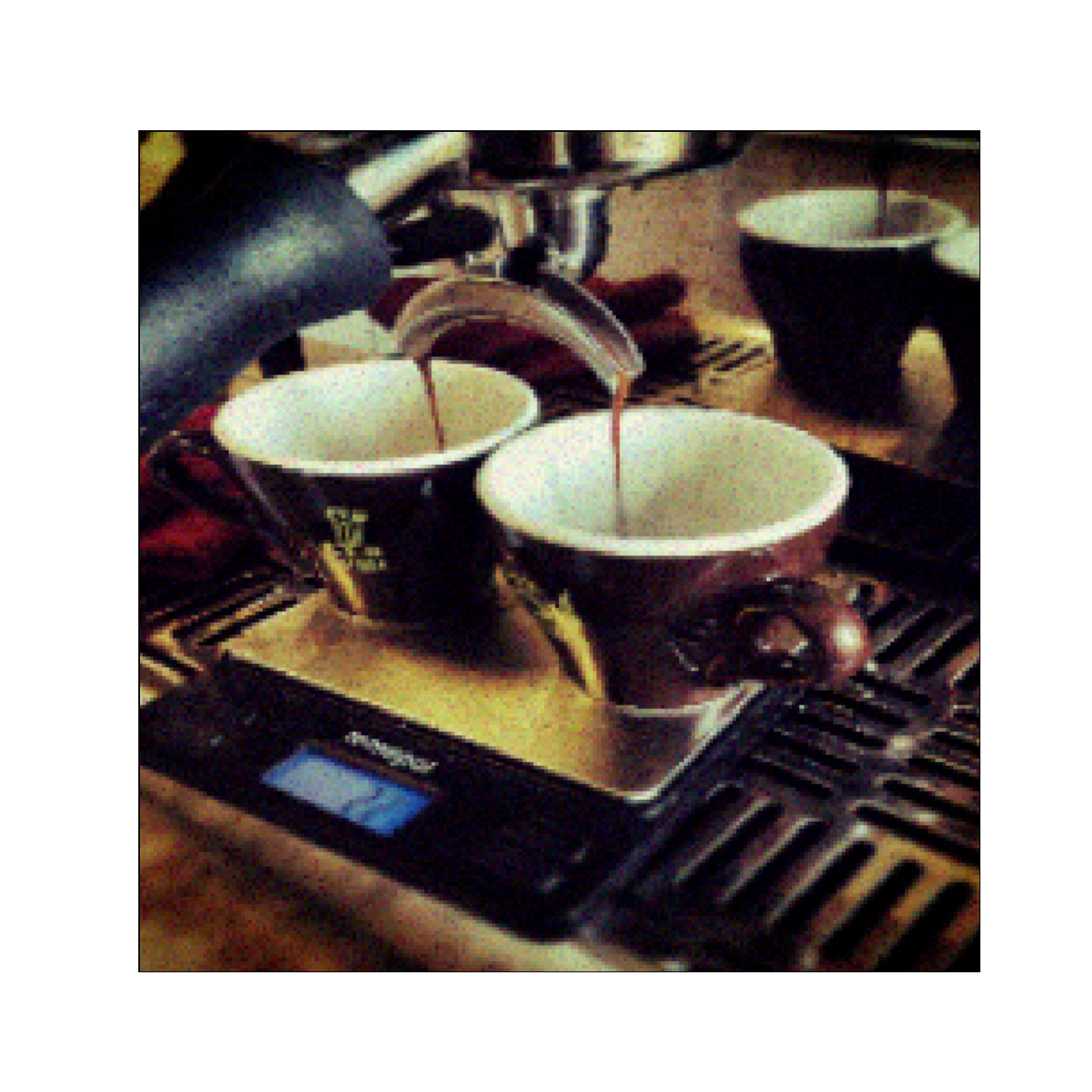}};
\node[below=\labeldist of row24.north] {\footnotesize{Typewriter}};
\node[right=\imdist of row24, xshift=\xshift] (row25) {\includegraphics[width=\signsize]{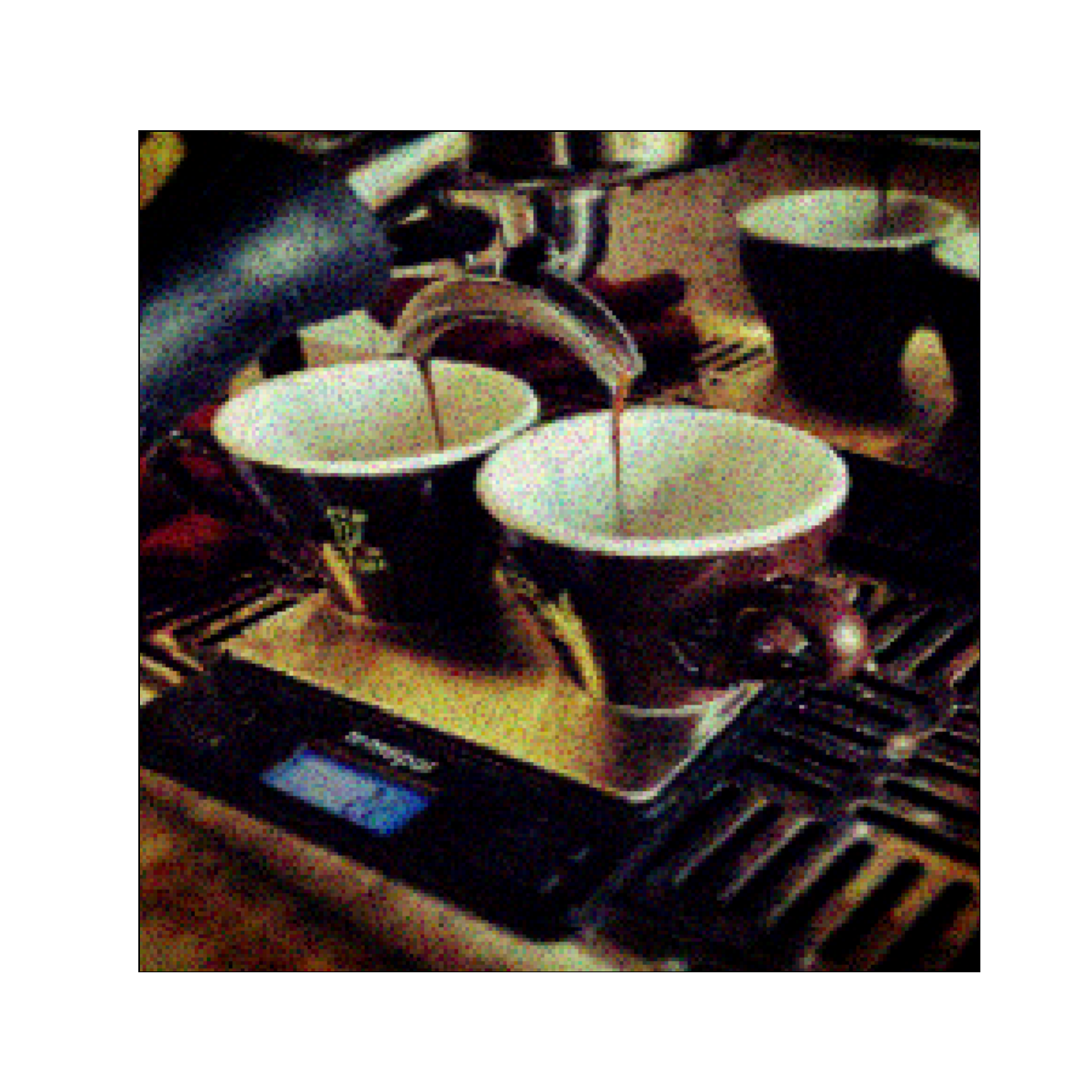}};
\node[below=\labeldist of row25.north] {\footnotesize{Typewriter}};
\node[right=\imdist of row25, xshift=\xshift] (row26) {\includegraphics[width=\signsize]{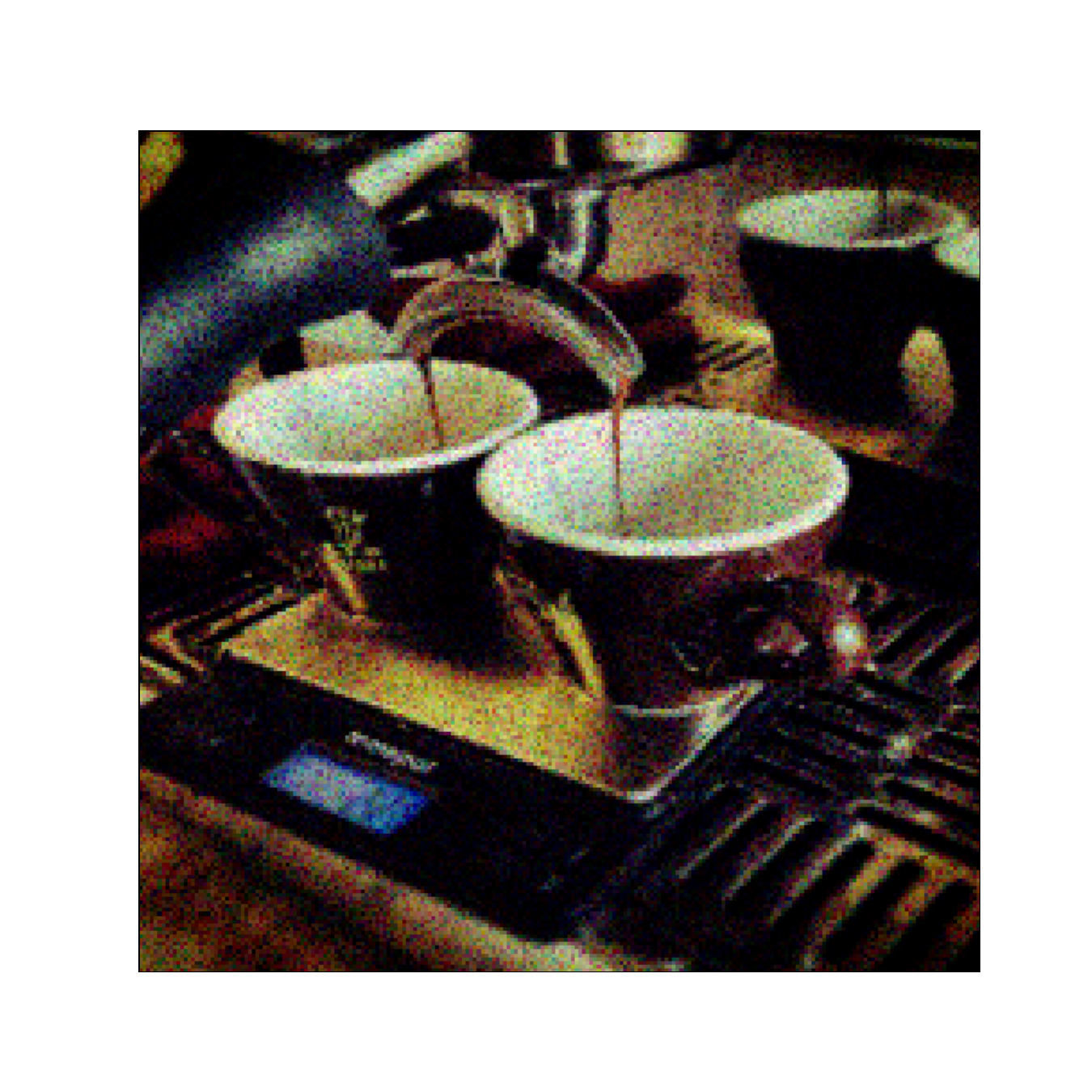}};
\node[below=\labeldist of row26.north] {\footnotesize{Typewriter}};

\end{tikzpicture}

%% file: figures/mia-adaptive.tex
\definecolor{blue}{HTML}{1f77b4}
\definecolor{orange}{HTML}{ff7f0e}
\definecolor{green}{HTML}{2ca02c}

\begin{tikzpicture}

  \centering
  \begin{axis}[
        ybar=4pt, axis on top,
        height=4cm, width=0.8\textwidth,
        bar width=0.3cm,
        %ymajorgrids,
        %major grid style={draw=white},
        enlarge x limits=0.25,
        ymin=0, ymax=100,
        axis x line*=bottom,
        %axis y line*=right,
        %y axis line style={opacity=0},
        %ytick= {50,100},
        nodes near coords,
        nodes near coords align={vertical},
        point meta=rawy,
        tick label style={font=\footnotesize},
        label style={font=\footnotesize},
        legend style={
            at={(0.5,-0.55)},
            anchor=north,
            legend columns=6,
            font=\footnotesize,
            /tikz/every even column/.append style={column sep=0.3cm}
        },
        legend entries={VGG (3x3), ResNet (3x3), Inception (3x3), VGG (5x5), ResNet (5x5), Inception (5x5)},
        ylabel={Success Rate (\%)},
        xlabel={\# Training points},
        %symbolic x coords={
        %   100,
        %   1000,
        %   2000},
        xtick={0,10,20,30},
        xticklabels = {100,1000,3000},
       xtick=data,
       tick pos=left,
       nodes near coords style={font=\footnotesize},
       legend image code/.code={
        \draw [#1] (0cm,-0.1cm) rectangle (0.2cm,0.25cm); },
    ]
    
    % VGG (3x3)
    \addplot [draw=none, fill=red, postaction={pattern=north east lines}] coordinates {
      (10, 56)
      (20, 75) 
      (30, 80)
      };

    % ResNet (3x3)
    \addplot [draw=none, fill=blue, postaction={pattern=north east lines}] coordinates {
      (10, 18)
      (20, 46) 
      (30, 51)
      };
      
    % Inception (3x3)
   \addplot [draw=none,fill=green, postaction={pattern=north east lines}] coordinates {
      (10, 10)
      (20, 32) 
      (30, 31)
      };

    % VGG (5x5)
    \addplot [draw=none, fill=red, postaction={pattern=horizontal lines}] coordinates {
      (10, 77)
      (20, 92) 
      (30, 94)
      };

    % ResNet (5x5)
    \addplot [draw=none, fill=blue, postaction={pattern=horizontal lines}] coordinates {
      (10, 50)
      (20, 69) 
      (30, 68)
      };

    % Inception (5x5)
   \addplot [draw=none,fill=green, postaction={pattern=horizontal lines}] coordinates {
      (10, 28)
      (20, 45) 
      (30, 48)
      };
      
      \addlegendimage{line legend,black,dashed}

    \coordinate (A) at (axis cs:100,90);
    \coordinate (O1) at (rel axis cs:0,0);
    \coordinate (O2) at (rel axis cs:1,0);

    %\draw [black,sharp plot,dashed] (A -| O1) -- (A -| O2);

    %\legend{SR ($L^0$), SR ($L^1$), $\mathcal{A} (L^0)$, $\mathcal{A} (L^0)$, clean}
  \end{axis}
  \end{tikzpicture}

%% file: figures/filters.tex
\begin{tikzpicture}

\node(11) {$\begin{bmatrix}
		-0.54 & 0.80 & -0.53\\
		0.80 & -1.00 & 0.80\\
		-0.51 & 0.81 & -0.53
	\end{bmatrix}$};

\node[above=of 11, yshift=-0.8cm] {\textbf{ResNet ($3\times 3$) , Channel 1}};
%\node[left=of 11, xshift=0.2cm,rotate=90,anchor=north] {\textbf{ResNet}};

\node[right=of 11, xshift=-0.2cm] (12) {$\begin{bmatrix}
		0.05 & 0.52 & -0.63\\
        0.52 & -1.00 & 0.6\\
        -0.58 & 0.55 & 0.02
	\end{bmatrix}$};

 \node[above=of 12, yshift=-0.8cm] {\textbf{VGG ($3\times 3$) , Channel 1}};

\node[right=of 12, xshift=-0.2cm] (13) {$\begin{bmatrix}
		0.04 & 0.29 & -0.37 & -0.43 & 0.24\\
0.27 & -0.6 & 0.23 & 0.34 & -0.26\\
-0.46 & 0.34 & 1.0 & 0.3 & -0.44\\
-0.34 & 0.45 & 0.24 & -0.68 & 0.23\\
0.26 & -0.58 & -0.35 & 0.43 & 0.02\\
	\end{bmatrix}$};

\node[above=of 13, yshift=-0.8cm] {\textbf{ResNet ($5\times 5$) , Channel 2}};
 
\end{tikzpicture}

%% file: figures/appendix-model-filters.tex
\begin{tikzpicture}[baseline,remember picture]

\newcommand{\signsize}{0.12\textwidth}
\newcommand{\imdist}{0.4cm}
\newcommand{\rowdist}{0.1em}
\newcommand{\labeldist}{2.2cm}

\node (11) {\includegraphics[width=\signsize]{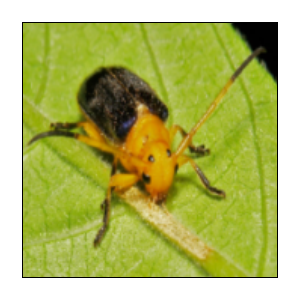}};
\node[left=0.4cm of 11.west, anchor=north, rotate=90]{\footnotesize{Input}};
\node[right=\imdist of 11] (12) {\includegraphics[width=\signsize]{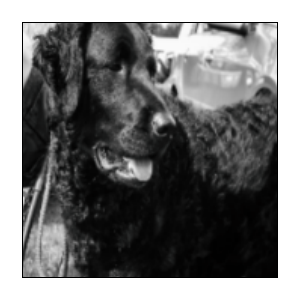}};
\node[right=\imdist of 12] (13) {\includegraphics[width=\signsize]{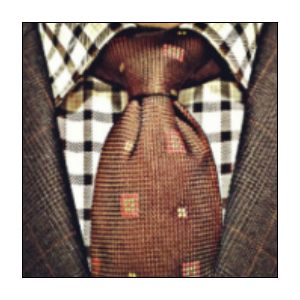}};
\node[right=\imdist of 13] (14) {\includegraphics[width=\signsize]{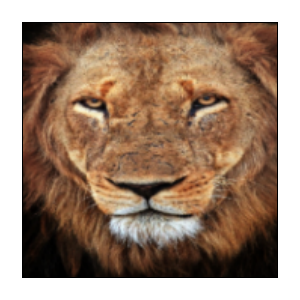}};
\node[right=\imdist of 14] (15) {\includegraphics[width=\signsize]{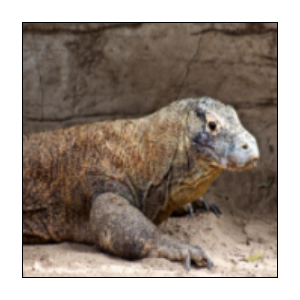}};
\node[right=\imdist of 15] (16) {\includegraphics[width=\signsize]{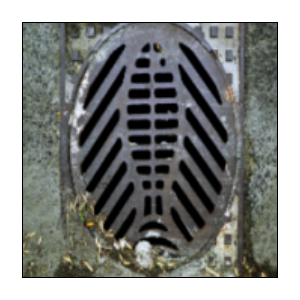}};

\node[below=\rowdist of 11.south] (21) {\includegraphics[width=\signsize]{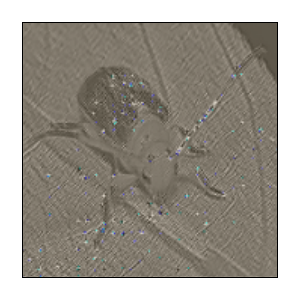}};
\node[left=0.4cm of 21.west, anchor=north, rotate=90]{\footnotesize{VGG}};
\node[right=\imdist of 21] (22) {\includegraphics[width=\signsize]{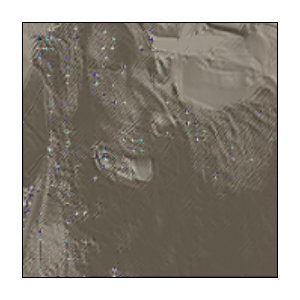}};
\node[right=\imdist of 22] (23) {\includegraphics[width=\signsize]{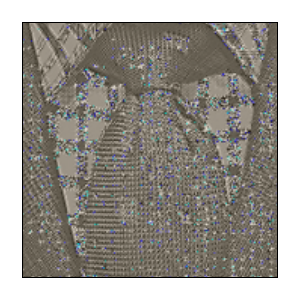}};
\node[right=\imdist of 23] (24) {\includegraphics[width=\signsize]{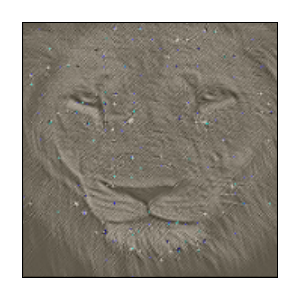}};
\node[right=\imdist of 24] (25) {\includegraphics[width=\signsize]{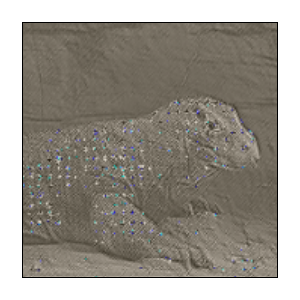}};
\node[right=\imdist of 25] (26) {\includegraphics[width=\signsize]{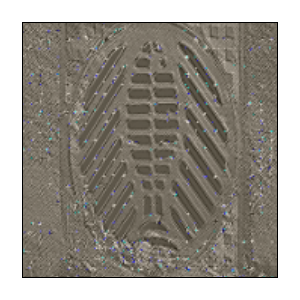}};

\node[below=\rowdist of 21.south] (31) {\includegraphics[width=\signsize]{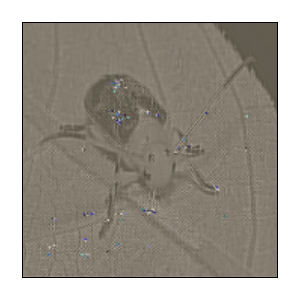}};
\node[left=0.4cm of 31.west, anchor=north, rotate=90]{\footnotesize{ResNet}};
\node[right=\imdist of 31] (32) {\includegraphics[width=\signsize]{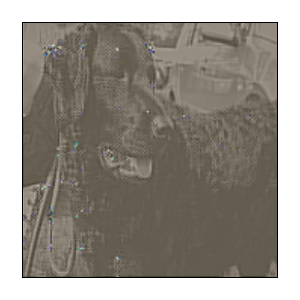}};
\node[right=\imdist of 32] (33) {\includegraphics[width=\signsize]{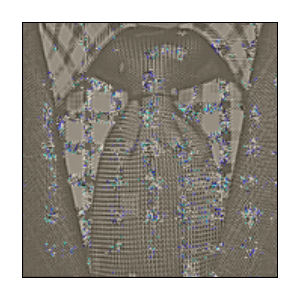}};
\node[right=\imdist of 33] (34) {\includegraphics[width=\signsize]{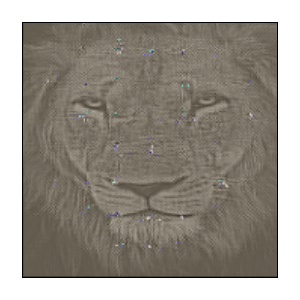}};
\node[right=\imdist of 34] (35) {\includegraphics[width=\signsize]{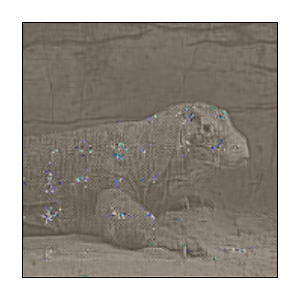}};
\node[right=\imdist of 35] (36) {\includegraphics[width=\signsize]{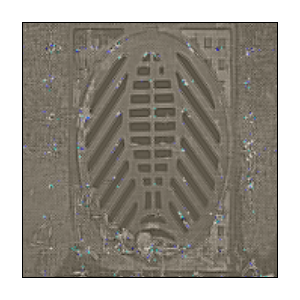}};

\node[below=\rowdist of 31.south] (41) {\includegraphics[width=\signsize]{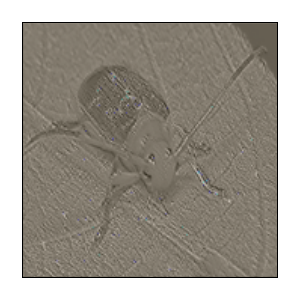}};
\node[left=0.4cm of 41.west, anchor=north, rotate=90]{\footnotesize{Inception}};
\node[right=\imdist of 41] (42) {\includegraphics[width=\signsize]{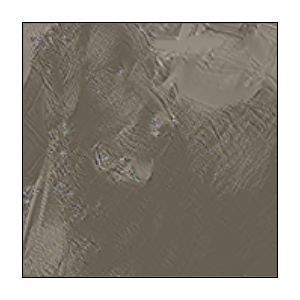}};
\node[right=\imdist of 42] (Roundabout-filter) {\includegraphics[width=\signsize]{figures/adversarial-filters/dog-resnet.pdf}};
\node[right=\imdist of 42] (43) {\includegraphics[width=\signsize]{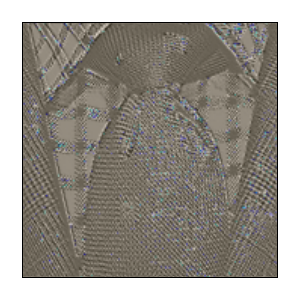}};
\node[right=\imdist of 43] (44) {\includegraphics[width=\signsize]{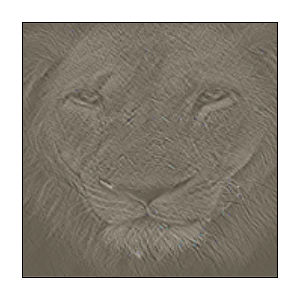}};
\node[right=\imdist of 44] (45) {\includegraphics[width=\signsize]{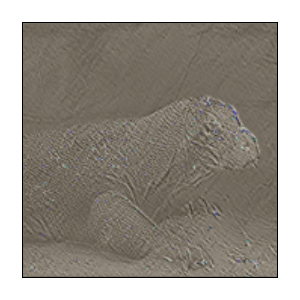}};
\node[right=\imdist of 45] (46) {\includegraphics[width=\signsize]{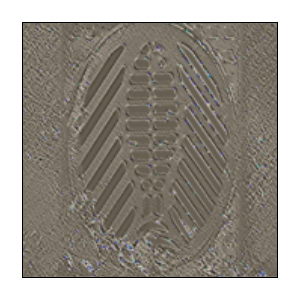}};

\end{tikzpicture}

%% file: figures/mia-adversarial-examples.tex
\begin{tikzpicture}[baseline,remember picture]

\newcommand{\signsize}{0.15\textwidth}
\newcommand{\imdist}{0.4cm}
\newcommand{\rowdist}{0.2cm}
\newcommand{\labeldist}{2.6cm}

\node (11) {\includegraphics[width=\signsize]{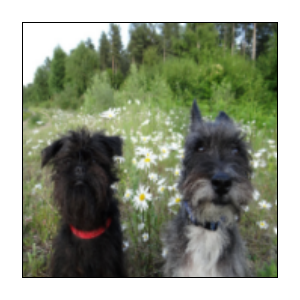}};
\node[left=0.4cm of 11.west, anchor=north, rotate=90]{\footnotesize{Input}};
\node[below=\labeldist of 11.north] {\footnotesize{Affenpinscher}};
\node[right=\imdist of 11] (12) {\includegraphics[width=\signsize]{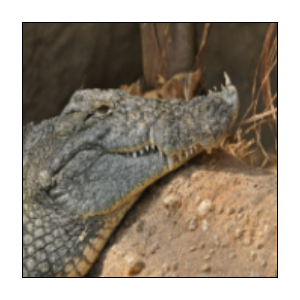}};
\node[below=\labeldist of 12.north] {\footnotesize{African crocodile}};
\node[right=\imdist of 12] (13) {\includegraphics[width=\signsize]{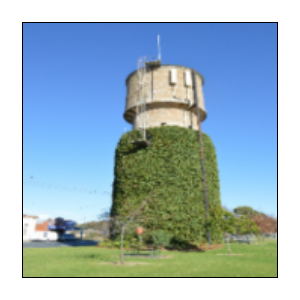}};
\node[below=\labeldist of 13.north] {\footnotesize{Castle}};
\node[right=\imdist of 13] (14) {\includegraphics[width=\signsize]{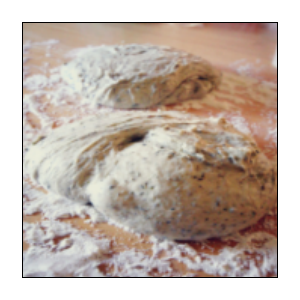}};
\node[below=\labeldist of 14.north] {\footnotesize{Dough}};
\node[right=\imdist of 14] (15) {\includegraphics[width=\signsize]{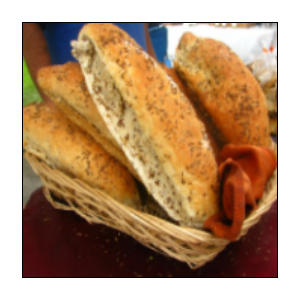}};
\node[below=\labeldist of 15.north] {\footnotesize{French loaf}};

\node[below=\rowdist of 11.south] (21) {\includegraphics[width=\signsize]{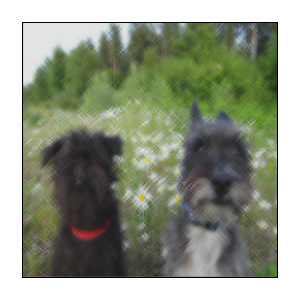}};
\node[left=0.4cm of 21.west, anchor=north, rotate=90]{\footnotesize{PSNR=20}};
\node[below=\labeldist of 21.north] {\footnotesize{Standard Schnauzer}};
\node[right=\imdist of 21] (22) {\includegraphics[width=\signsize]{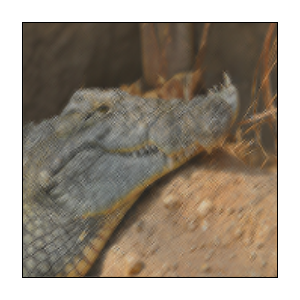}};
\node[below=\labeldist of 22.north] {\footnotesize{Window Screen}};
\node[right=\imdist of 22] (23) {\includegraphics[width=\signsize]{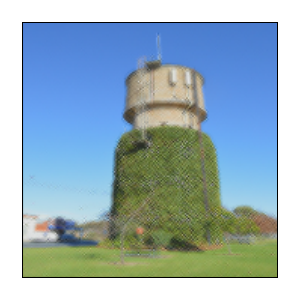}};
\node[below=\labeldist of 23.north] {\footnotesize{Water bottle}};
\node[right=\imdist of 23] (24) {\includegraphics[width=\signsize]{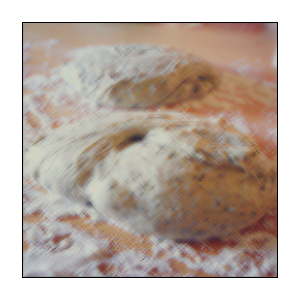}};
\node[below=\labeldist of 24.north] {\footnotesize{Clumber}};
\node[right=\imdist of 24] (25) {\includegraphics[width=\signsize]{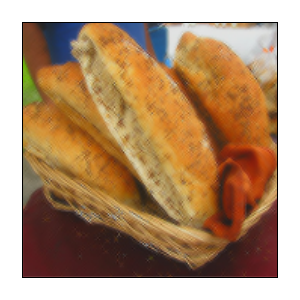}};
\node[below=\labeldist of 25.north] {\footnotesize{Ear}};

\node[below=\rowdist of 21.south] (31) {\includegraphics[width=\signsize]{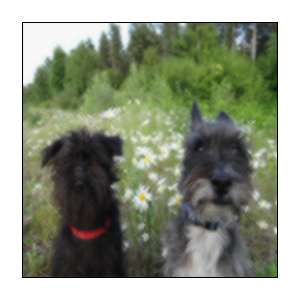}};
\node[left=0.4cm of 31.west, anchor=north, rotate=90]{\footnotesize{PSNR=30}};
\node[below=\labeldist of 31.north] {\footnotesize{Standard Schnauzer}};
\node[right=\imdist of 31] (32) {\includegraphics[width=\signsize]{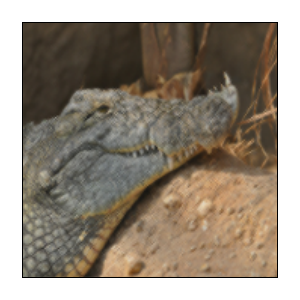}};
\node[below=\labeldist of 32.north] {\footnotesize{Window Screen}};
\node[right=\imdist of 32] (33) {\includegraphics[width=\signsize]{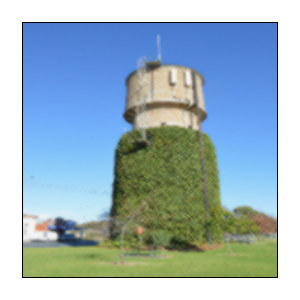}};
\node[below=\labeldist of 33.north] {\footnotesize{Water bottle}};
\node[right=\imdist of 33] (34) {\includegraphics[width=\signsize]{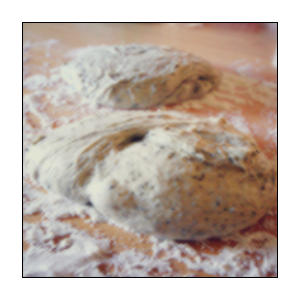}};
\node[below=\labeldist of 34.north] {\footnotesize{English setter}};
\node[right=\imdist of 34] (35) {\includegraphics[width=\signsize]{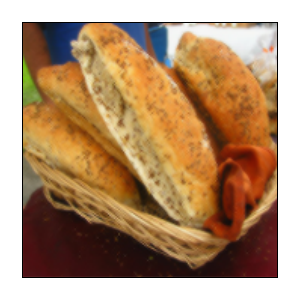}};
\node[below=\labeldist of 35.north] {\footnotesize{Boa constrictor}};

\end{tikzpicture}

%% file: figures/filters-appendix.tex
\begin{tikzpicture}

\node(11) {$\begin{bmatrix}
		0.05 & 0.52 & -0.63\\
        0.52 & -1.00 & 0.6\\
        -0.58 & 0.55 & 0.02
	\end{bmatrix}$};

\node[above=of 11, yshift=-0.8cm] {\textbf{VGG}};
\node[left=of 11, xshift=0.2cm,rotate=90,anchor=north] {\textbf{Channel 1}};

\node[right=of 11, xshift=-0.2cm] (12) {$\begin{bmatrix}
		-0.54 & 0.80 & -0.53\\
        0.80 & -1.00 & 0.80\\
        -0.51 & 0.81 & -0.53
	\end{bmatrix}$};

 \node[above=of 12, yshift=-0.8cm] {\textbf{ResNet}};

\node[right=of 12, xshift=-0.2cm] (13) {$\begin{bmatrix}
		-0.56 & 0.32 & 0.45\\
        0.29 & -1.00 & 0.15\\
        0.44 & 0.45 & -0.51
	\end{bmatrix}$};

\node[above=of 13, yshift=-0.8cm] {\textbf{Inception}};

\node[below=of 11] (21) {$\begin{bmatrix}
		-0.94 & 0.82 & -0.39\\
        0.86 & -0.62 & 0.94\\
        -0.4 & 0.87 & -1.00
	\end{bmatrix}$};
\node[left=of 21, xshift=0.2cm,rotate=90,anchor=north] {\textbf{Channel 2}};

\node[right=of 21, xshift=-0.2cm] (22) {$\begin{bmatrix}
		-0.62 & 0.92 & -0.69\\
        1.00 & -0.99 & 0.96\\
        -0.7 & 0.95 & -0.64
	\end{bmatrix}$};

\node[right=of 22, xshift=-0.2cm] (23) {$\begin{bmatrix}
		-0.62 & 0.22 & 0.09\\
        0.69 & -1.00 & 0.33\\
        0.5 & 0.58 & -0.5
	\end{bmatrix}$};

\node[below=of 21] (31) {$\begin{bmatrix}
		0.05 & 0.52 & -0.51\\
        0.52 & -1.00 & 0.47\\
        -0.52 & 0.44 & 0.07
	\end{bmatrix}$};
\node[left=of 31, xshift=0.2cm,rotate=90,anchor=north] {\textbf{Channel 3}};

\node[right=of 31, xshift=-0.2cm] (32) {$\begin{bmatrix}
		-0.41 & 0.67 & -0.41\\
        0.63 & -1.00 & 0.67\\
        -0.39 & 0.64 & -0.39
	\end{bmatrix}$};

\node[right=of 32, xshift=-0.2cm] (33) {$\begin{bmatrix}
		-0.3 & 0.61 & 0.51\\
        -0.12 & -1.00 & 0.37\\
        0.28 & 0.11 & -0.43
	\end{bmatrix}$};
 
\end{tikzpicture}

%% file: figures/filters-appendix-5x5.tex
\begin{tikzpicture}

\node(11) {$\begin{bmatrix}
		-0.97 & 0.34 & 0.63 & -0.05 & -0.23\\
0.53 & -0.95 & 0.03 & 0.5 & -0.09\\
0.71 & 0.0 & -0.45 & 0.12 & 0.56\\
-0.38 & 0.6 & 0.36 & -1.0 & 0.33\\
-0.04 & -0.11 & 0.38 & 0.27 & -0.8\\
	\end{bmatrix}$};

\node[above=of 11, yshift=-0.8cm] {\textbf{Channel 1}};

\node[right=of 11, xshift=-0.2cm] (12) {$\begin{bmatrix}
		0.04 & 0.29 & -0.37 & -0.43 & 0.24\\
0.27 & -0.6 & 0.23 & 0.34 & -0.26\\
-0.46 & 0.34 & 1.0 & 0.3 & -0.44\\
-0.34 & 0.45 & 0.24 & -0.68 & 0.23\\
0.26 & -0.58 & -0.35 & 0.43 & 0.02\\
	\end{bmatrix}$};

\node[above=of 12, yshift=-0.8cm] {\textbf{Channel 2}};

\node[below=of 11, yshift=-0.2cm] (13) {$\begin{bmatrix}
		-0.96 & 0.18 & 0.89 & 0.16 & -0.32\\
0.26 & -0.83 & -0.13 & 0.44 & 0.06\\
0.99 & -0.28 & -1.0 & 0.07 & 0.8\\
-0.03 & 0.55 & -0.03 & -0.94 & 0.31\\
-0.29 & 0.22 & 0.77 & 0.15 & -0.89\\
	\end{bmatrix}$};

\node[above=of 13, yshift=-0.8cm] {\textbf{Channel 3}};
 
\end{tikzpicture}